%% file: arxiv.tex
\documentclass[10pt,twocolumn,letterpaper]{article}

\usepackage[pagenumbers]{cvpr} %

\input{preamble}

\definecolor{cvprblue}{rgb}{0.21,0.49,0.74}
\usepackage[pagebackref,breaklinks,colorlinks,allcolors=cvprblue]{hyperref}

\usepackage{times}
\usepackage{epsfig}
\usepackage{graphicx}
\usepackage{url}            %
\usepackage{booktabs}       %
\usepackage{amsfonts}       %
\usepackage{nicefrac}       %
\usepackage{microtype}      %
\usepackage{color}
\usepackage{colortbl}
\usepackage{dsfont}
\usepackage{algorithm,algpseudocode}
\usepackage{algorithmicx}
\usepackage{amsmath,amssymb}
\usepackage{multirow}
\usepackage{makecell}
\usepackage{tabularx}
\newcounter{abs}

\usepackage{mathtools}
\usepackage{amsthm}
\usepackage{caption}
\algblock{ParFor}{EndParFor}
\algnewcommand\algorithmicparfor{\textbf{parfor}}
\algnewcommand\algorithmicpardo{\textbf{do}}
\algnewcommand\algorithmicendparfor{\textbf{end\ parfor}}
\algrenewtext{ParFor}[1]{\algorithmicparfor\ #1\ \algorithmicpardo}
\algrenewtext{EndParFor}{\algorithmicendparfor}

\newtheorem{theorem}{Theorem}

\usepackage{hyperref}

\definecolor{Green}{rgb}{0.85882353, 0.90980392, 0.84705882}
\definecolor{Blue}{rgb}{0.80392157, 0.91372549, 0.97254902}
\definecolor{Purple}{rgb}{0.9019608 0.9019608, 0.9803922}
\definecolor{RPurple}{rgb}{0.92941176 0.787451, 0.7880392}
\definecolor{Grey}{rgb}{0.81176471, 0.81176471, 0.81176471}
\definecolor{DGreen}{rgb}{0.15882353, 0.80980392, 0.14705882}
\newcommand{\customfootnotetext}[2]{{%
\renewcommand{\thefootnote}{#1}%
\footnotetext[0]{#2}}}%

\title{Parallel Sequence Modeling via Generalized Spatial Propagation Network}

\author{Hongjun Wang\textsuperscript{1,2,$\dagger$}, Wonmin Byeon$^1$, Jiarui Xu$^3$, Jinwei Gu$^1$, Ka Chun Cheung$^1$, \\ Xiaolong Wang\textsuperscript{1,3}, Kai Han$^2$, Jan Kautz$^1$, Sifei Liu$^1$
\\
$\textsuperscript{\rm 1}$NVIDIA,\quad$\textsuperscript{\rm 2}$The University of Hong Kong, \quad$\textsuperscript{\rm 3}$University of California, San Diego
}

\begin{document}
\maketitle
\customfootnotetext{}{$\dagger$ Hongjun Wang was an intern at NVIDIA during the project.}
\input{sec/0_abstract}    
\input{sec/1_intro}
\input{sec/2_rw}
\input{sec/formulation}
\input{sec/3_method}
\input{sec/4_exp}

\input{sec/5_con}

\clearpage
\input{sec/X_supp}
\clearpage
{
    \small
    \bibliographystyle{ieeenat_fullname}
    \bibliography{main}
}

\end{document}

%% file: preamble.tex
\usepackage[dvipsnames]{xcolor}

%% file: sec/0_abstract.tex
\begin{abstract}
We present the Generalized Spatial Propagation Network (GSPN), a new attention mechanism optimized for vision tasks that inherently captures 2D spatial structures. Existing attention models, including transformers, linear attention, and state-space models like Mamba, process multi-dimensional data as 1D sequences, compromising spatial coherence and efficiency. GSPN overcomes these limitations by directly operating on spatially coherent image data and forming dense pairwise connections through a line-scan approach. Central to GSPN is the Stability-Context Condition, which ensures stable, long-context propagation across 2D sequences and reduces the effective sequence length to $\sqrt{N}$ for a square map with N elements, which significantly enhances computational efficiency. With learnable, input-dependent weights and no reliance on positional embeddings, GSPN achieves superior spatial fidelity and state-of-the-art performance in vision tasks, including ImageNet classification, class-guided image generation, and text-to-image generation. Notably, GSPN accelerates SD-XL with softmax-attention by over $84\times$ when generating 16K images. Project page: \url{https://whj363636.github.io/GSPN/}
\end{abstract}

%% file: sec/1_intro.tex
\section{Introduction}

Transformers have revolutionized machine learning, driving significant advances in fields such as natural language processing~\cite{brown2020language,kocon2023chatgpt,touvron2023llama} and computer vision~\cite{ViT16x16,touvron2021training,liu2021swin,carion2020end,Peebles2023DiT}. Their attention mechanisms—self-attention for in-depth context modeling and cross-attention for multi-source integration—offer exceptional flexibility in capturing intricate dependencies across data elements. Despite their impact, transformers face two core limitations. First, their quadratic computational complexity hampers efficiency at large scales, making them computationally intensive, especially when processing long context sources such as high-resolution images. Second, transformers treat data as structure-agnostic tokens that overlook the spatial coherence. This disregard for spatial structure diminishes their suitability for vision tasks, where maintaining positional relationships is crucial.

To address the efficiency limitations of transformers, recent approaches have aimed to reduce computational complexity from quadratic to linear with respect to sequence length. Linear attention methods~\cite{rabe2021self,peng2021random,han2023flatten} either leverage kernel methods~\cite{choromanski2020rethinking,peng2021random}, or reorder computations to exploit the associative property of matrix multiplication~\cite{katharopoulos2020transformers,lu2021soft}. For instance, by changing the computation order from the standard $(QK^\top)V$ to $Q(K^\top V)$, the overall computational complexity can be reduced from $\mathcal{O}(N^2)$ to $\mathcal{O}(N)$. Alternatively, State-Space Models (SSMs), such as Mamba~\cite{gu2023mamba}, leverage linear recurrent dynamics to capture long-range dependencies efficiently with $\mathcal{O}(N)$ complexity. However, both methods abstract away spatial structure, a crucial element for vision tasks that rely on spatial coherence.

\begin{figure}[t]
    \centering
    \includegraphics[width=\linewidth]{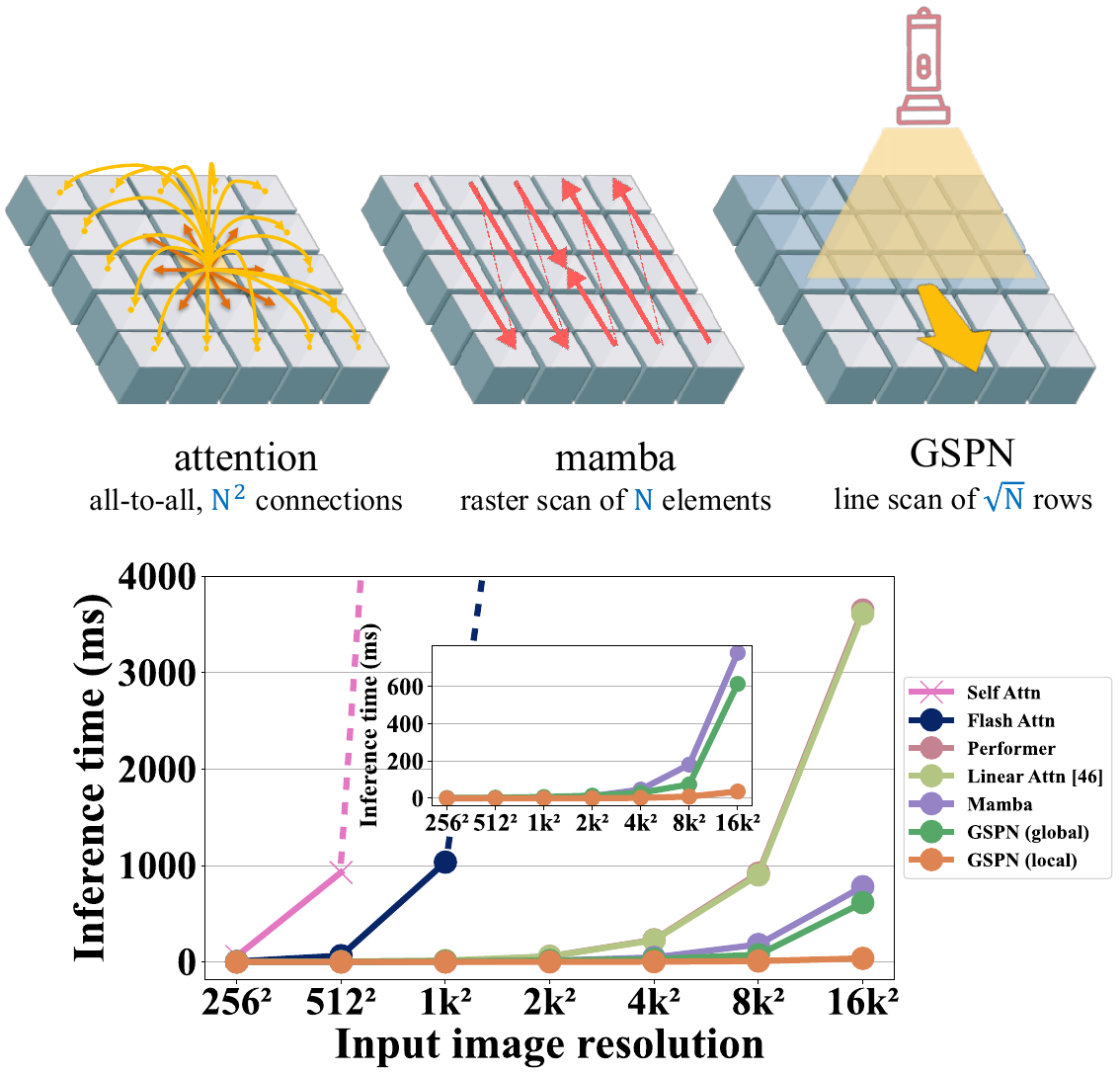}
    \caption{\textbf{Comparison of attention mechanisms and computational efficiency.} \textbf{Top:} Softmax attention (left), mamba (middle), and our GSPN (right).  \textbf{Bottom:} Inference speed comparison across different input sizes on A100 GPU, demonstrating GSPN's superior efficiency for high-resolution input. Dashed lines indicate quadratic extrapolation due to memory explosion. Global and local GSPN are detailed in~\Cref{para:glob_locl}. 
    }
    \label{fig:teaser}
\end{figure}
One way to enhance spatial coherence is to extend 1D \textit{raster scan} to 2D \textit{line scan}, i.e., to perform linear propagation across rows and columns~\cite{van2016pixel,liu2017learning}. However, 2D linear propagation presents significant challenges: In 2D, scalar weights become matrices that link each pixel to its neighbors of the previous row or column, leading to accumulated matrix multiplications during propagation. If these matrices have large eigenvalues, instability may arise with exponential growth; if too small, signals may decay quickly, causing information to vanish~\cite{gu2023mamba,liu2017learning}. Reducing the weights to maintain stability limits the receptive field and weakens long-distance dependencies. Balancing stability with a large receptive field remains challenging for 2D linear propagation.

In this work, we introduce the Generalized Spatial Propagation Network (GSPN), a linear attention mechanism optimized for multi-dimensional data such as images. Central to GSPN is the Stability-Context Condition, which ensures both stability and effective long-range context propagation across 2D sequences by maintaining a consistent propagation weight norm. This condition allows information from distant elements to influence large spatial areas meaningfully while preventing exponential growth in dependencies, thus enabling stable and context-aware propagation essential for vision tasks. With a linear line-scan operation, GSPN parallelizes propagation across rows and columns, reducing the effective sequence length to $\sqrt{N}$, significantly enhancing the computational efficiency, as illustrated in~\Cref{fig:teaser}. This makes GSPN a robust and scalable framework that overcomes the key limitations of existing attention mechanisms by inherently capturing 2D spatial structures.

During propagation, GSPN computes a weighted sum for each pixel using pixels from its previous row or column, with weights that are learnable and input-dependent. Retaining key designs from~\cite{liu2017learning}, GSPN uses a 3-way connection for parameter efficiency, while a 4-direction integration ensures full pixel connectivity, thereby forming dense pairwise connections through the line-scan manner. We present two variants of GSPN: one that captures global context across the entire input and another that focuses on local regions for faster propagation. These variants allow GSPN to seamlessly integrate into modern vision architectures as a drop-in replacement for existing attention modules. In addition, we introduce a learnable merger that aggregates spatial information from all scanning directions, which enhances the model's ability to adapt dynamically to the 2D structure of visual data. Specifically, by inherently incorporating positional information through scanning, GSPN eliminates the need for positional embeddings and avoids common aliasing issues~\cite{yang2023emernerf,darcet2024vision}.

As a new sub-quadratic attention block tailored for vision, it is crucial to comprehensively benchmark GSPN to showcase its effectiveness and efficiency. To this end, we conduct extensive evaluations across a diverse range of visual tasks, including deterministic tasks like ImageNet classification, and generative tasks such as class-conditional generation (DiT) and text-to-image (T2I) generation. Our results are compelling: GSPN achieves competitive performance compared to CNNs and Transformers in recognition tasks, and surpasses SoTA diffusion transformer-based models in class-conditional generation while using only $65.6\%$ of the parameters. In text-to-image generation, GSPN demonstrates its sublinear scaling with pixel count, matching original performance on high-resolution tasks (e.g., 16K) while reducing inference time by up to $84\times$ on the SD-XL model. These findings position GSPN as a powerful, efficient alternative to traditional transformers for 2D vision tasks.

%% file: sec/2_rw.tex
\section{Related work}
\noindent\textbf{Vision Transformers.} The emergence of transformer architectures challenges the traditional dominance of CNNs~\cite{krizhevsky2012imagenet,he2016deep,liu2022convnet} in computer vision. Vision Transformer (ViT)~\cite{ViT16x16} pioneers this paradigm shift by processing images as sequences of embedded patches, catalyzing numerous architectural innovations~\cite{carion2020end,liu2021swin,liu2021swin,touvron2021training,caron2021emerging}. Due to the versatility of attention from different modalities, Transformer-based models are also widely employed in multimodal tasks~\cite{hertz2022prompttoprompt,li2023blip2}. However, their quadratic computational complexity in attention mechanisms poses significant challenges for processing long sequences, limiting their broader application to high-resolution or long-context tasks.

\noindent\textbf{Efficient Vision Transformers.} The quadratic computational complexity of the attention module catalyzes diverse optimization strategies in the field. Early approaches introduce sparse attention mechanisms~\cite{child2019generating,wu2020lite,kitaev2020reformer,wang2021pyramid,zhu2023biformer,lingle2023transformer}. To reduce the overall computational complexity $\mathcal{O}(N^2)$ for computing $(QK^\top)V$, parallel developments also explore architectural modifications, including local attention mechanisms~\cite{ho2019axial}, hierarchical processing~\cite{liu2021swin,graham2021levit}, hybrid architectures~\cite{Wu_2021_ICCV,dai2021coatnet}. Subsequent works focus on linear attention formulations~\cite{katharopoulos2020transformers,choromanski2020rethinking,wang2020linformer,xiong2021nystromformer,hua2022transformer}. These linear approaches encompass channel-wise attention methods with $\mathcal{O}(ND^2)$ complexity~\cite{shen2021efficient,ali2021xcit}, kernel-based approximations of softmax using exponential functions~\cite{katharopoulos2020transformers,lu2021soft,peng2021random} or Taylor expansions~\cite{banerjee2020exploring}, and matrix decomposition techniques~\cite{xiong2021nystromformer,wang2020linformer}.

\noindent\textbf{Sequence Modeling in 1D and 2D Space.} Recurrent neural networks (RNNs) such as LSTMs~\cite{hochreiter1997long}, GRUs~\cite{chung2014empirical}, 2D-LSTMs~\cite{graves2007multi,byeon2015scene} have been widely adopted for sequential modeling through non-linear transformations. Their sequential modeling inherently limits computational efficiency and scalability. Additionally, their non-linear nature often compromises long-term dependency modeling due to gradient vanishing and exploding issues~\cite{hochreiter1991untersuchungen,pascanu2013difficulty}, where distant past information either fails to influence future states or causes optimization instability. 
Recent state space models (SSM)~\cite{fu2023hungry,gu2023mamba} address these limitations through linear propagation, ensuring stability and long-range context modeling. Several works~\cite{nguyen2022s4nd,baron20232,zhu2024ViM,liu2024vmamba,li2024mamba} adapt SSM for image processing by flattening 2D data into 1D sequences. Although avoiding the direct application of non-linearities in propagation, these approaches potentially sacrifice inherent spatial structure information.

\noindent\textbf{Spatial Propagation Networks.} Spatial Propagation Network (SPN)~\cite{liu2017learning} first introduces linear propagation for 2D data, initially targeting sparse-to-dense prediction tasks like segmentation, as a single-layer module atop deep CNNs. However, the potential of scaling SPN as a foundational architecture like ViT remains unexplored and its sequential processing across different directions limits computational efficiency. Additionally, SPN does not address long-range propagation, which is essential for foundational blocks in high-level tasks. Our GSPN network advances these efforts by implementing parallel row/column-wise propagation, enabling efficient affinity matrix learning while maintaining gradient stability and long-range correlation. We provide both theoretical analysis and empirical results to demonstrate GSPN's effectiveness as a compelling alternative to ViT and Mamba architectures.

%% file: sec/formulation.tex
\begin{figure}[t]
    \centering
    \includegraphics[width=0.98\linewidth]{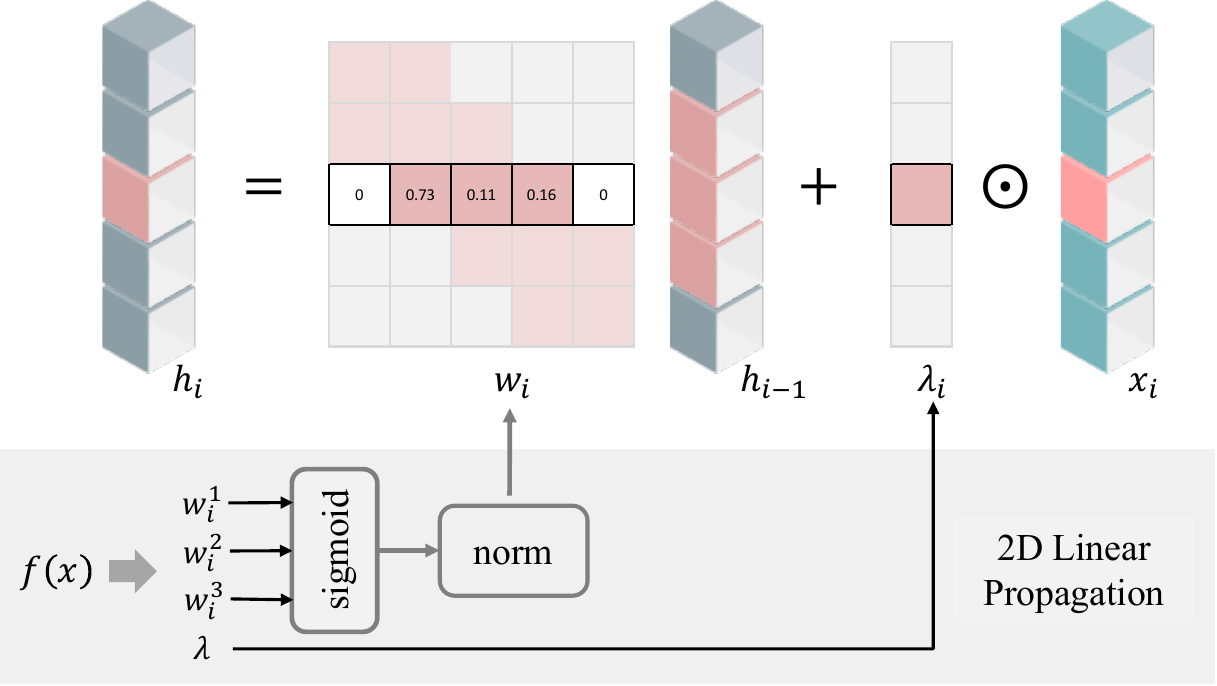}
    \caption{\textbf{2D Linear Propagation.} For the $i$-th row, each item in the hidden state $h_i$ is computed by: (1) a weighted sum of three neighboring values from the hidden layer $h_{i-1}$, where weights form a normalized tridiagonal matrix $w_i$, and (2) the element-wise product of the current input $x_i$ with $\lambda$. $w_i$ and $\lambda$ are both learnable and input-dependent parameters. The weights in $w_i$ are obtained by applying sigmoid activation followed by row-wise normalization.}\label{fig:prop}
    \label{fig:2dprog}
\end{figure}
\section{Method}
In this section, we introduce the formulation of 2D linear propagation (Section~\ref{sec:formulation}), establish the mathematical foundation for ensuring stable and long-range context propagation (Section~\ref{sec:condition}), and detail the key design elements of GSPN (Section~\ref{sec:key}).

\subsection{2D Linear Propagation} \label{sec:formulation}
Linear propagation in 2D proceeds through sequential row-by-row or column-by-column processing. Consider a 2D image $x \in \mathbb{R}^{n \times n \times C}$ (we assume a square image for simplicity, though the approach generalizes to arbitrary dimensions). All following examples and presentations use row-by-row propagation as the illustrative cases. The 2D propagation follows a linear recurrent process:
\begin{equation}
    h_i^c = w_i^c h_{i-1}^c + \lambda_i^c \odot x_i^c, \quad i \in [1, n-1], \; c \in [0, C-1]
    \label{eq:2d}
\end{equation}
where $h \in \mathbb{R}^{n \times n \times C}$ is the hidden layer. Here, $x_i$ and $h_i$ denote the $i$-th row of the input $x$ and hidden state $h$. For each channel $c$, the propagation uses two learnable parameters: $w_i^c \in \mathbb{R}^{n \times n}$, an $n \times n$ matrix that weights $h_{i-1}^c$, and $\lambda_i^c \in \mathbb{R}^{n}$, which scales $x_i^c$ element-wise using $\odot$. Omitting the channel index $c$ for simplicity, we apply an output layer element-wise via $u_i \in \mathbb{R}^{n}$:
\vspace{-1mm}
\begin{equation}\vspace{-3mm}
    y_i = u_i \odot h_i
    \label{eq:out}
\end{equation}

\noindent\textbf{Expanded Form.} We denote the vectorized sequence of concatenated rows of hidden states and inputs as $H = \langle h_1^T, h_2^T, \dots, h_n^T \rangle^T$ and $X = \langle x_1^T, x_2^T, \dots, x_n^T \rangle^T$. Extending Eq.~\eqref{eq:2d}, we have $H = GX$, where $G$ is a lower triangular $N \times N$ matrix with $n \times n$ sub-matrices relating $h_i$ and $x_j$, defined as:
\begin{equation}
    G_{ij} = \begin{cases} 
        \prod_{\tau=j+1}^{i} w_\tau \lambda_j, & j \in [0, i-1] \\
        \lambda_j, & i = j 
    \end{cases}
    \label{eq:G}
\end{equation}
Leveraging Eq.~\eqref{eq:out}, the output $y_i$ can be represented as a weighted sum of $X$:
\vspace{-2mm}
\begin{equation}\vspace{-2mm}
    y_i=u_i\sum_{j=0}^{t}\prod_{\tau=j+1}^{i}w_\tau\lambda_jx_j
    \label{eq:outex}
\end{equation}
Here, we slightly abuse notation by using $\lambda_j$ and $u_i$ to denote $n \times n$ diagonal matrices with their original vector values on the main diagonal.

\noindent\textbf{Relation to Linear Attention.} By substituting $x_j$ with values $V_j$ and parameterizing $u_i$ and $\lambda_i$ with feed-forward network layers, i.e., $u_i = f_Q(x_i)$ and $\lambda_i = f_K(x_i)$, analogous to query and key representations, we can rewrite $y_i$ as:
\vspace{-2mm} \begin{equation}\vspace{-2mm} y_i = f_Q(x_i) \sum_{j=0}^{i} \prod_{\tau=j+1}^{i} w_\tau f_K(x_j) V_j \label{eq:linear_attn} \end{equation}
Intuitively, Eq.~\eqref{eq:linear_attn} represents a non-normalized linear attention mechanism with causal masking \cite{katharopoulos2020transformers}, where the additional propagation matrix $\prod_{\tau=j+1}^{i} w_\tau$ modulates the strength of attention.

\subsection{Stability-Context Condition} \label{sec:condition}
We investigate how to design $\prod_{\tau=j+1}^{i} w_\tau$ to achieve stability and effective long-range propagation. Letting $W_{ij} = \prod_{\tau=j+1}^{i} w_\tau$, dense interactions between $h_i$ and $x'_j = \lambda_j x_j$ can be ensured, even when $i$ and $j$ are far apart, if (i) $W_{ij}$ is a dense matrix, and (ii) $\sum_{j=0}^{n-1} W_{ij} = 1$, so that each element in $h_i$ is a weighted average of all elements in $x'_j$. In the following, we introduce Theorems \ref{the:1} and \ref{the:2}, collectively referred to as the \textit{Stability-Context Condition}, which meet these requirements.
\begin{theorem} \label{the:1}
    If all the matrices $w_{\tau}$ are row stochastic, then $\sum_{j=0}^{n-1} W_{ij} = 1$ is satisfied.
\end{theorem}

\noindent \textbf{Definition.} A matrix $T$ is row stochastic if (i) all elements are non-negative, $T_{ij} \geq 0$ for all $(i,j)$; and (ii) the sum of elements in each row is 1, $\sum_{j} T_{ij} = 1$ for all $i$.

\noindent \textbf{Proof.} The theorem holds because the product of row stochastic matrices is also row stochastic. See the Appendix for the complete proof.

\begin{theorem}\label{the:2}
    The stability of Eq.~\eqref{eq:2d} is ensured when all matrices $w_{\tau}$ are row stochastic.
\end{theorem}
\noindent \textbf{Proof.} Making $w_\tau$ row stochastic is a sufficient condition to ensure stability presented in \cite{liu2017learning}. See the Appendix for the complete proof.

\subsection{Key Implementations for Propagation Layer}\label{sec:key}
A straightforward way to satisfy the \textit{Stability-Context Condition} is to learn a full matrix $w_\tau$ that outputs $n$ weights per pixel, i.e., connecting all pixels in the previous row to each pixel in the current row, and normalizing the weights so that they sum to 1. However, this approach significantly increases the number of feature dimensions. To address this, we adopt the key designs from \cite{liu2017learning}, where each pixel connects to three pixels from the previous row: the top-left, top-middle, and top-right pixels in the top-to-bottom propagation direction. As a result, $w_\tau$ becomes a tridiagonal matrix. Importantly, multiplying multiple tridiagonal matrices results in a dense $W_{ij}$, satisfying the requirements outlined in Section \ref{sec:condition}. In addition, we adopt the line scan from 4 directions, i.e., left-to-right, top-to-bottom, and verse-visa, to ensure dense pairwise connections among all pixels.

For each propagation direction, to ensure the matrix $w_{\tau}$ to be row stochastic, let $w_{\tau, i, j}$ be an element in row $i$ and column $j$. Suppose each row $i$ has $m_i$ non-zero entries. As illustrated in~\Cref{fig:2dprog}, $w_i^k$ is the pre-sigmoid value of the $k$-th non-zero element in row $i$. We apply the sigmoid function to each non-zero element $\sigma(w_i^k) = 1/(1 + e^{-w_i^k})$ and then normalize the entries in row $i$ so they sum to 1. Thus, each non-zero element in row $i$ can be expressed as:
\begin{equation}
w_{\tau, i, k} = \frac{\sigma(w_i^k)}{\sum_{k'=1}^{m_i} \sigma(w_i^{k'})},
\end{equation}
where $k$ indexes the $m_i$ non-zero entries in the $i$-row.

\noindent\textbf{Efficient CUDA implementation.} 
We implement the linear propagation layer in Eq. \eqref{eq:2d} via a customized CUDA kernel. Our kernel function employs a parallelized structure with $p=512$ threads per block and $q = \frac{BCn^2 k (p + 1)}{p}$ blocks per grid, where $B$ represents the mini-batch size and $k$ denotes the number of propagation directions. Each thread processes a single pixel in the input image along the propagation direction, enabling full parallelization across the batch, channels, and rows/columns orthogonal to the propagation. This design effectively reduces the kernel loop length to $n$, facilitating efficient and scalable linear propagation. See the Appendix for more details.

%% file: sec/3_method.tex
\section{GSPN Architecture}
\begin{figure*}[t]
    \centering
    \includegraphics[width=0.95\linewidth]{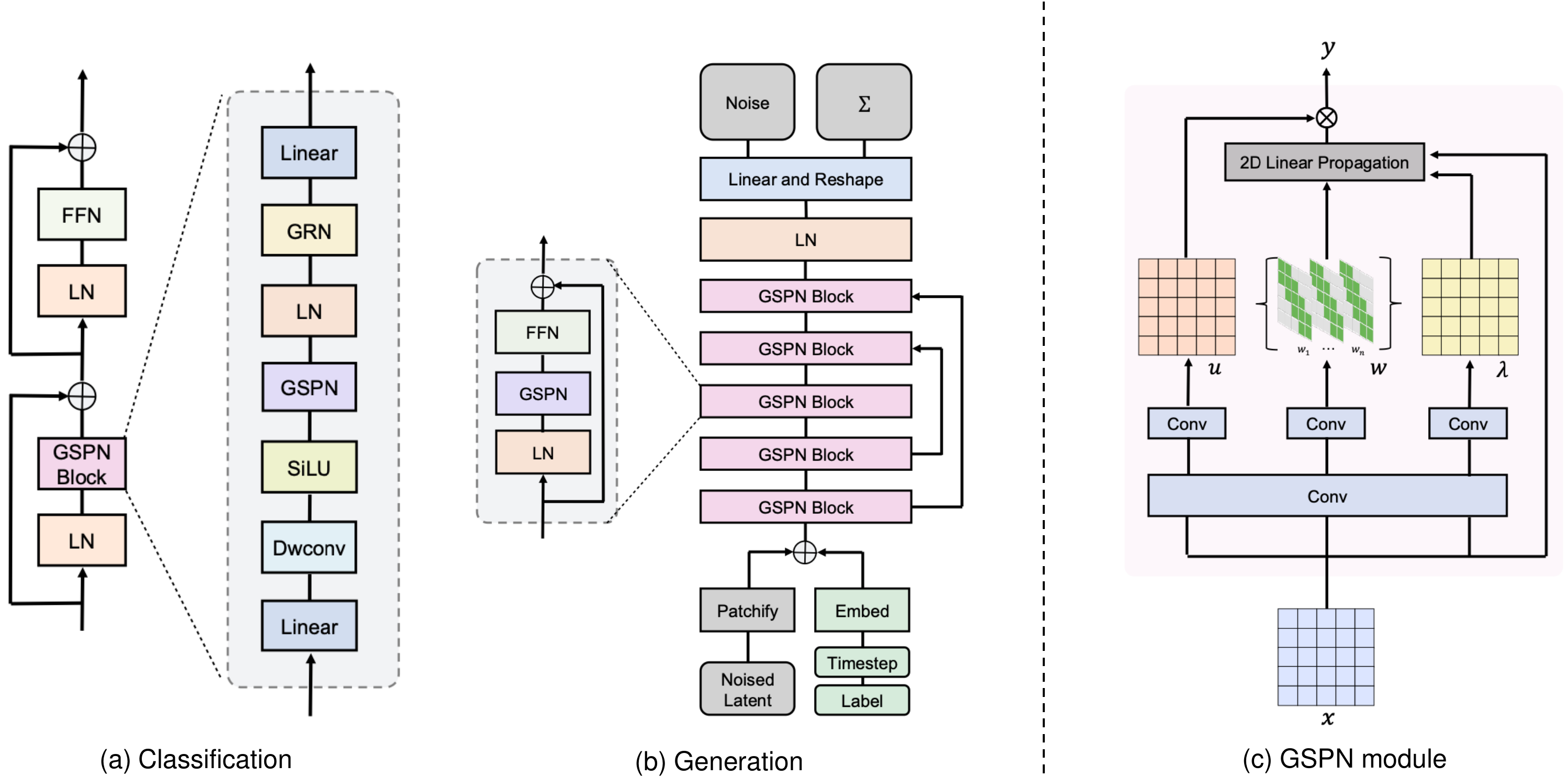}
    \caption{\textbf{Architecture of GSPN.} We design different GSPN blocks for discriminative and generative task. Both blocks share the same GSPN module with our core 2D Linear Propagation mechanism illustrated in~\Cref{fig:2dprog}.}
    \label{fig:arch}
\end{figure*}

GSPN is a generic sequence propagation module that can be seamlessly integrated into neural networks for various visual tasks. This section outlines the macro designs of the GSPN block for discriminative tasks (\Cref{fig:arch}(a)) and generative task (\Cref{fig:arch} (b)), built on the core GSPN module shown in \Cref{fig:arch} (c). Following that, we share key insights gained from extensive empirical trials that guided the development and refinement of GSPN.

\subsection{GSPN module Macro Design}
As illustrated in~\Cref{fig:arch} (a) and (b), our top-level designs for both image classification and generation tasks share fundamental architectural principles, drawing from successful vision models~\cite{wang2021pyramid,liu2021swin,liu2022convnet}. To make a fair comparison, both architectures only integrate commonly used blocks including separable convolutions~\cite{chollet2017xception}, layer normalization (LN)~\cite{ba2016layernorm}, GRN~\cite{woo2023convnext}, feed-forward networks (FFN)~\cite{vaswani2017attention}, and non-linear activations~\cite{hendrycks2016gelu,elfwing2018sigmoid}. 

\noindent\textbf{Local \vs Global GSPN.} 
\label{para:glob_locl}
The GSPN described earlier operates across entire sequences to capture long-range dependencies, which we refer to as global GSPN. To enhance efficiency, we introduce local GSPN, which reduces the propagation sequence length by restricting it to localized regions.
Local GSPN divides one spatial dimension into $g$ non-overlapping groups, where each group $\mathcal{G}_k = \{i: i_k \leq i < i_{k+1}\}$ contains a subset of indices satisfying $\bigcup_{k=1}^{g} \mathcal{G}_k = [1, n-1]$ and $\mathcal{G}_j \cap \mathcal{G}_k = \emptyset$ for $j \neq k$. Within each group, local GSPN computes hidden states according to Eq.~\eqref{eq:2d}: $h_i^c = w_i^c h_{i-1}^c + \lambda_i^c \odot x_i^c$ for $i \in \mathcal{G}_k, c \in [0, C-1]$. This grouping strategy enables parallel computation, reducing complexity by a factor of $g$ compared to global GSPN, achieving $\mathcal{O}(1)$ complexity in the extreme case where $g = n$. We adopt a default group size $g=2$ for local GSPN. For image classification tasks that prioritize semantic understanding, we employ more global GSPN modules to capture long-range dependencies and holistic features. In contrast, for dense prediction and generation tasks that require fine-grained spatial details and local consistency, we predominantly utilize local GSPN modules to preserve spatial structure and local coherence.

\noindent\textbf{GSPN Module.} As shown in~\Cref{fig:arch} (c), we apply a shared $1\times1$ convolutions for dimensionality reduction, followed by three separate $1\times1$ convolutions to generate input-dependent parameters $u$, $w$ and $\lambda$ for 2D linear propagation. These projections and 2D linear propagation are encapsulated within a modular GSPN unit, designed to integrate seamlessly across the architectures presented in the following.

\begin{figure*}[t]
    \centering
    \includegraphics[width=\linewidth]{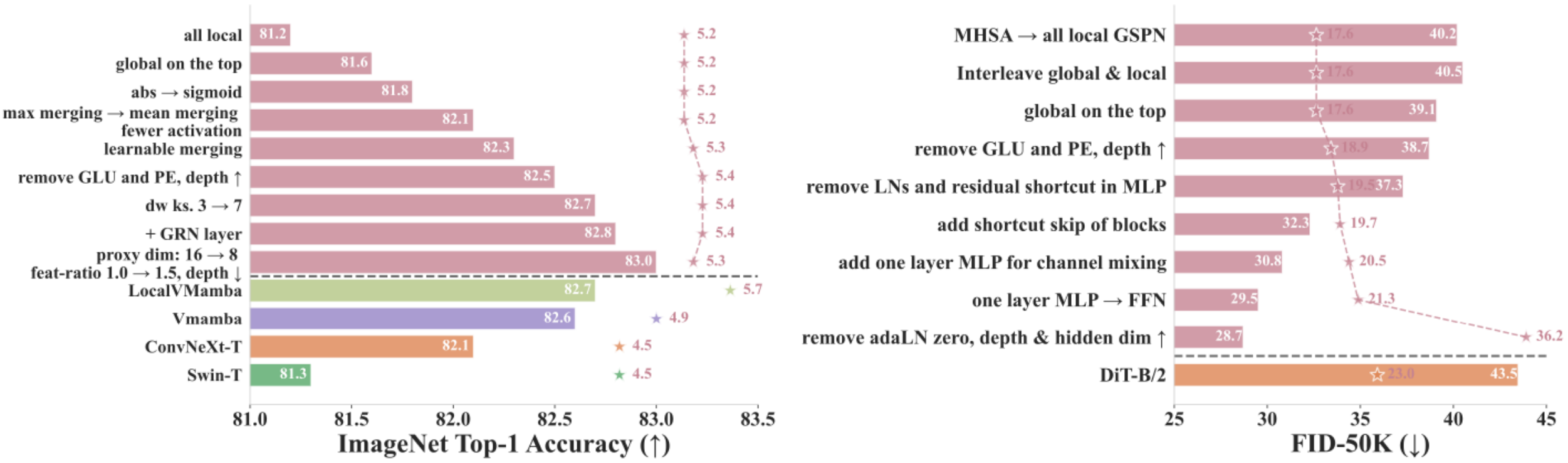}
    \caption{\textbf{Design choice of GSPN with benchmark methods in classification accuracy and computational efficiency for both classification and generation tasks.} 
    A higher Top-1 accuracy and a lower FID indicate better model capability. Our GSPN can outperform other methods under similar FLOPs.}
    \label{fig:ablation}
    \vspace{-3pt}
\end{figure*}

\noindent\textbf{Image Classification Module.} Following~\cite{liu2021swin,meng2021sdedit}, we implement a four-level hierarchical architecture for image classification by stacking well-designed GSPN blocks (\Cref{fig:arch} (a)) as direct substitution of self-attention with GSPN modules yields suboptimal results (top 2 bars in~\Cref{fig:ablation} left). Between adjacent two levels, each downsampling operation halves the spatial dimensions. We employ local GSPN blocks in levels 1-2 for efficient processing at higher resolutions, while utilizing global GSPN blocks in levels 3-4 for contextual integration at lower resolutions. This architectural design balances computational efficiency with representational capacity, enabling effective local feature extraction at higher resolutions while facilitating global information aggregation at deeper layers through the transition from local to global GSPN blocks. 

\noindent\textbf{Class-conditional Image Generation Module.} Analogous to classification task, we redesign the architecture for generation, since direct replacement of self-attention with GSPN modules yields limited improvements (top 3 bars in~\Cref{fig:ablation} right). \Cref{fig:arch} (b) provides an overview of the GSPN architecture for class-conditional image generation networks. The model integrates timestep $t$ and conditional information $y$ through vector embedding addition instead of AdaIN-zero~\cite{Peebles2023DiT}. The architecture features skip connections that concatenate shallow and deep layer representations, followed by linear projections. Compared with self-attention module, we notably remove positional embeddings and incorporate a FFN for channel mixing. The final decoding stage transforms the sequence of hidden states through layer normalization and linear projection, reconstructing the spatial layout to predict both noise and diagonal covariance with dimensions matching the input. 

\noindent\textbf{Text-to-Image Generation Module.} Text-to-image (T2I) tasks require extensive training data, making it impractical to train from scratch. Instead of adopting the aforementioned image generation module shown in \Cref{fig:arch} (b), we integrate GSPN module in \Cref{fig:arch} (c) directly into the Stable Diffusion (SD) architecture by replacing all self-attention layers with GSPN modules. To leverage prior knowledge and accelerate training, we initialize the $u$, $\lambda$, and $x$ parameters in Eq.\eqref{eq:outex} using the pre-trained query, key, and value weight matrices from SD, capitalizing on the mathematical relationship between GSPN and linear attention as shown in Eq.\eqref{eq:linear_attn}. Detailed training procedures are provided in~\Cref{subsec:impl}.

\input{tables/cls}

\subsection{Principle of GSPN design}\label{sec:principle}
We outline key design principles for GSPN at both macro and micro scales, contrasting them with classic attention and Mamba modules. These principles, derived from empirical analysis, serve as guidelines for optimizing GSPN networks across various computer vision tasks.

\noindent\textbf{Combination of global and local GSPN.} Local GSPN blocks are applied to the early stages for efficient processing of fine-grained spatial details. The subsequent global GSPN blocks aggregate long-range contextual information for higher-level semantic understanding. This hierarchical design achieves an optimal trade-off between accuracy and efficiency, and improves both tasks by 0.4\% accuracy and 1.1 FID as shown in~\Cref{fig:ablation}. 

\noindent\textbf{Learnable merging better than manual design.} In previous work, SPN~\cite{liu2017learning} applies manual merging via max pooling operations to combine multi-directional scan information, while GSPN implements a linear layer to dynamically aggregate features from different scanning directions. This data-driven merging strategy enables the network to adaptively weight and combine directional information based on the current input propagation (see ``learnable merging" in~\Cref{fig:ablation} for the comparison with merging approaches including max and mean).

\noindent\textbf{No Positional Embedding.} Our GSPN desgin demonstrates that explicit positional embeddings are unnecessary for both classification and generation tasks, as spatial information is inherently encoded through the scanning process. This design choice effectively addresses the aliasing issues noted in recent works~\cite{yang2023emernerf,darcet2024vision}, while departing from conventional approaches in DiT-based methods that rely on learnable APE with sinusoidal functions~\cite{vaswani2017attention} or RoPE~\cite{su2024roformer}. 
As shown in~\Cref{fig:ablation} (\ie, ``remove PE"), GSPN can remove PE for both tasks without negative effects.

\noindent\textbf{Fewer normalization layers.} By minimizing normalization layers, we achieve improved computational efficiency and reduced model complexity without sacrificing performance (\Cref{fig:ablation}). This suggests that the traditional extensive use of normalization layers may be redundant in GSPN as the $w$ is normalized through the stability-context condition.

\noindent\textbf{GLU is not effective in GSPN.} Unlike Mamba~\cite{gu2023mamba}, empirical evidence (\ie, ``remove GLU" in \Cref{fig:ablation}) indicates that Gated Linear Units (GLU) do not provide significant benefits to GSPN. This observation suggests that 2D linear propagation may already provide gating mechanisms.

%% file: tables/cls.tex
\begin{table*}[h!]
\renewcommand{\arraystretch}{1.3}
    \caption{
    \textbf{Performance of models on ImageNet at the resolution of $224^2$.} Colors denote different backbone types: \textbf{\textcolor{Blue}{blue}} for ConvNets, \textbf{\textcolor{Purple}{purple}} for Transformers, and \textbf{\textcolor{RPurple}{magenta}} for raster scan (i.e., 1D linear propagation) methods.
    }
    \label{tab:cls-imagenet}
    \centering
    \begin{minipage}{.33\textwidth}
    \scriptsize
    \centering
    \setlength{\tabcolsep}{.6pt}
    \begin{tabular}{l | c | c | c c }
    \toprule
    \multirow{3}{*}{\makecell[c]{Model}}    & \multirow{3}{*}{\makecell[c]{Backbone}}     & \multirow{3}{*}{\makecell[c]{Param \\ (M)}}   & \multicolumn{2}{c}{IN-1K} \\
    \cline{4-5}
    ~ & ~ & ~ &  \multirow{2}{*}{\makecell[c]{MAC \\ (G)}} & \multirow{2}{*}{\makecell[c]{Acc \\ (\%)}} \\ 
    ~ & ~ & ~ & ~ & ~ \\
    \midrule
    \rowcolor{Blue}
    ConvNeXT-T \cite{liu2022convnet} &  ConvNet   & 29 & 4.5 & 82.1  \\
    \rowcolor{Blue}
    MambaOut-Tiny \cite{yu2024mambaout} & ConvNet & 27 & 4.5 & 82.7 \\
    \rowcolor{Purple}
    Swin-T \cite{liu2021swin} &  Transformer   & 29 & 4.5 & 81.3 \\
    \rowcolor{Purple}
    CSWin-T \cite{dong2022cswin} &  Transformer  & 23 & 4.3 & 82.7  \\
    \rowcolor{Purple}
    CoAtNet-0 \cite{dai2021coatnet} &   Transformer  & 25 & 4.2 & 81.6 \\
    \rowcolor{RPurple}
    Vim-S \cite{zhu2024ViM} & Raster & 26 & 5.1 & 80.5 \\
    \rowcolor{RPurple}
    VMamba-T \cite{liu2024vmamba} &  Raster & 22 & 5.6 & 82.2 \\
    \rowcolor{RPurple}
    Mamba-2D-S \cite{li2024mamba} &  Raster & 24 & -- & 81.7 \\
    \rowcolor{RPurple}
    LocalVMamba-T \cite{huang2024localmamba} & Raster & 26 & 5.7 & 82.7 \\
    \rowcolor{RPurple}
    VRWKV-S \cite{duan2024visionrwkv} & Raster & 24 & 4.6 & 80.1 \\
    \rowcolor{RPurple}
    ViL-S \cite{alkin2024visionlstm} & Raster & 23 & 5.1 & 81.5 \\
    \rowcolor{RPurple}
    MambaVision-T \cite{hatamizadeh2024mambavision} & Raster & 32 & 4.4 & 82.3 \\
    \midrule
    \rowcolor{Green}
    \textbf{GSPN-T (Ours)} & Line & 30 & 5.3 & \textbf{83.0} \\
    \bottomrule
    \end{tabular}
    \end{minipage}
    \begin{minipage}{.33\textwidth}
    \scriptsize
    \centering
    \setlength{\tabcolsep}{.6pt}
    \begin{tabular}{l | c | c | c c }
    \toprule
    \multirow{3}{*}{\makecell[c]{Model}}    & \multirow{3}{*}{\makecell[c]{Backbone}}     & \multirow{3}{*}{\makecell[c]{Param \\ (M)}}   & \multicolumn{2}{c}{IN-1K} \\
    \cline{4-5}
    ~ & ~ & ~ &  \multirow{2}{*}{\makecell[c]{MAC \\ (G)}} & \multirow{2}{*}{\makecell[c]{Acc \\ (\%)}} \\ 
    ~ & ~ & ~ & ~ & ~ \\
    \midrule
    \rowcolor{Blue}
    ConvNeXT-S \cite{liu2022convnet} & ConvNet & 50 & 8.7 & 83.1 \\
    \rowcolor{Blue}
    MambaOut-Small \cite{yu2024mambaout} & ConvNet & 48 & 9.0 & \textbf{84.1} \\
    \rowcolor{Purple}
    T2T-ViT-19 \cite{yuan2021tokens} & Transformer & 39 & 8.5 & 81.9 \\
    \rowcolor{Purple}
    Focal-Small \cite{yang2022focal} & Transformer & 51 & 9.1 & 83.5 \\
    \rowcolor{Purple}
    NextViT-B \cite{li2022next} & Transformer & 45 & 8.3 & 83.2 \\
    \rowcolor{Purple}
    Twins-B \cite{chu2021twins} & Transformer & 56 & 8.3 & 83.1 \\
    \rowcolor{Purple}
    Swin-S \cite{liu2021swin} & Transformer & 50 & 8.7 & 83.0 \\
    \rowcolor{Purple}
    CoAtNet-1 \cite{dai2021coatnet} &  Transformer & 42 & 8.4 & 83.3 \\
    \rowcolor{Purple}
    UniFormer-B \cite{li2022uniformer} &  Transformer & 50 & 8.3 & 83.9 \\
    \rowcolor{RPurple}
    VMamba-S \cite{liu2024vmamba} &  Raster & 44 & 11.2 & 83.5 \\
    \rowcolor{RPurple}
    LocalVMamba-S \cite{huang2024localmamba} &  Raster & 50 & 11.4 & 83.7 \\
    \rowcolor{RPurple}
    MambaVision-S \cite{hatamizadeh2024mambavision} & Raster & 50 & 7.5 & 83.3 \\
    \midrule
    \rowcolor{Green}
    \textbf{GSPN-S (Ours)} & Line & 50 & 9.0 & 83.8 \\
    \bottomrule
    \end{tabular}
    \end{minipage}
    \begin{minipage}{.33\textwidth}
    \scriptsize
    \centering
    \setlength{\tabcolsep}{.6pt}
    \begin{tabular}{l | c | c | c c }
    \toprule
    \multirow{3}{*}{\makecell[c]{Model}}    & \multirow{3}{*}{\makecell[c]{Backbone}}     & \multirow{3}{*}{\makecell[c]{Param \\ (M)}}   & \multicolumn{2}{c}{IN-1K} \\
    \cline{4-5}
    ~ & ~ & ~ &  \multirow{2}{*}{\makecell[c]{MAC \\ (G)}} & \multirow{2}{*}{\makecell[c]{Acc \\ (\%)}} \\ 
    ~ & ~ & ~ & ~ & ~ \\
    \midrule
    \rowcolor{Blue}
    ConvNeXT-B \cite{liu2022convnet} & ConvNet & 89 & 15.4 & 83.8 \\
    \rowcolor{Blue}
    MambaOut-Base \cite{yu2024mambaout} & ConvNet & 85 & 15.8 & 84.2 \\
    \rowcolor{Purple}
    DeiT-B \cite{touvron2021training} & Transformer & 86 & 17.5 & 81.8 \\
    \rowcolor{Purple}
    Swin-B \cite{liu2021swin} & Transformer & 88 & 15.4 & 83.5 \\
    \rowcolor{Purple}
    CSwin-B \cite{dong2022cswin} & Transformer & 78 & 15.0 & 84.2 \\
    \rowcolor{Purple}
    CoAtNet-2 \cite{dai2021coatnet} &  Transformer & 75 & 15.7 & 84.1 \\
    \rowcolor{RPurple}
    Vim-B \cite{zhu2024ViM} & Raster & 98 & 17.5 & 81.9 \\
    \rowcolor{RPurple}
    VMamba-B \cite{liu2024vmamba} &  Raster & 89 & 15.4 & 83.9 \\
    \rowcolor{RPurple}
    Mamba-2D-B \cite{li2024mamba} &  Raster & 92 & -- & 83.0 \\
    \rowcolor{RPurple}
    VRWKV-B \cite{duan2024visionrwkv} & Raster & 94 & 18.2 & 82.0 \\
    \rowcolor{RPurple}
    ViL-B \cite{alkin2024visionlstm} & Raster & 89 & 18.6 & 82.4 \\
    \rowcolor{RPurple}
    MambaVision-B \cite{hatamizadeh2024mambavision} & Raster & 98 & 15.0 & 84.2 \\
    \midrule
    \rowcolor{Green}
    \textbf{GSPN-B (Ours)} & Line & 89 & 15.9 & \textbf{84.3} \\
    \bottomrule
    \end{tabular}
    \end{minipage}
\end{table*}

%% file: sec/4_exp.tex
\section{Experiment}
We evaluate GSPN across discriminative tasks (\Cref{sec:exp_cls}), class-conditional generation (\Cref{sec:exp_gen}), and text-to-image synthesis (\Cref{sec:exp_t2i}). Our experiments demonstrate GSPN's effectiveness as an efficient alternative to transformer-based architectures in both discriminative and generative domains. We further conduct ablation studies on architectural components and present qualitative results.

\subsection{Setup}

\textbf{Datasets.} For both image classification and class-conditional generation, we use ImageNet~\cite{deng2009imagenet} for training and evaluating. For text-to-image synthesis, we utilize a curated subset of 169k high-quality images from LAION~\cite{schuhmann2022laion}. We evaluate our model on MS-COCO~\cite{lin2014microsoft} validation set at $1024\times 1024$ resolution, each paired with five descriptive captions.

\label{subsec:impl}
\noindent\textbf{Implementation Details.} For image classification, we adopt the training protocol from~\cite{liu2021swin,liu2024vmamba}. For class-conditional generation, we follow DiT's training setup~\cite{Peebles2023DiT} but use a larger batch size (1024) and fewer epochs (1000) with cosine learning rate scheduling to accelerate training.
For text-to-image, to reduce the resources required for training, we adopt a knowledge distillation approach following~\cite{kim2023bk,liu2024linfusion}. The distillation process combines three loss components: the MSE loss between teacher and student noise predictions, the MSE loss between actual and predicted noise, and the accumulated MSE losses comparing teacher and student predictions after each block. We set both distillation hyper-parameters to 0.5. We use 512 and 1024 resolution for training SD-v1.5 and SD-XL respectively.
All the experiments are conducted on NVIDIA A100 GPUs.

\begin{figure*}[!t]
    \centering
    \includegraphics[width=\linewidth]{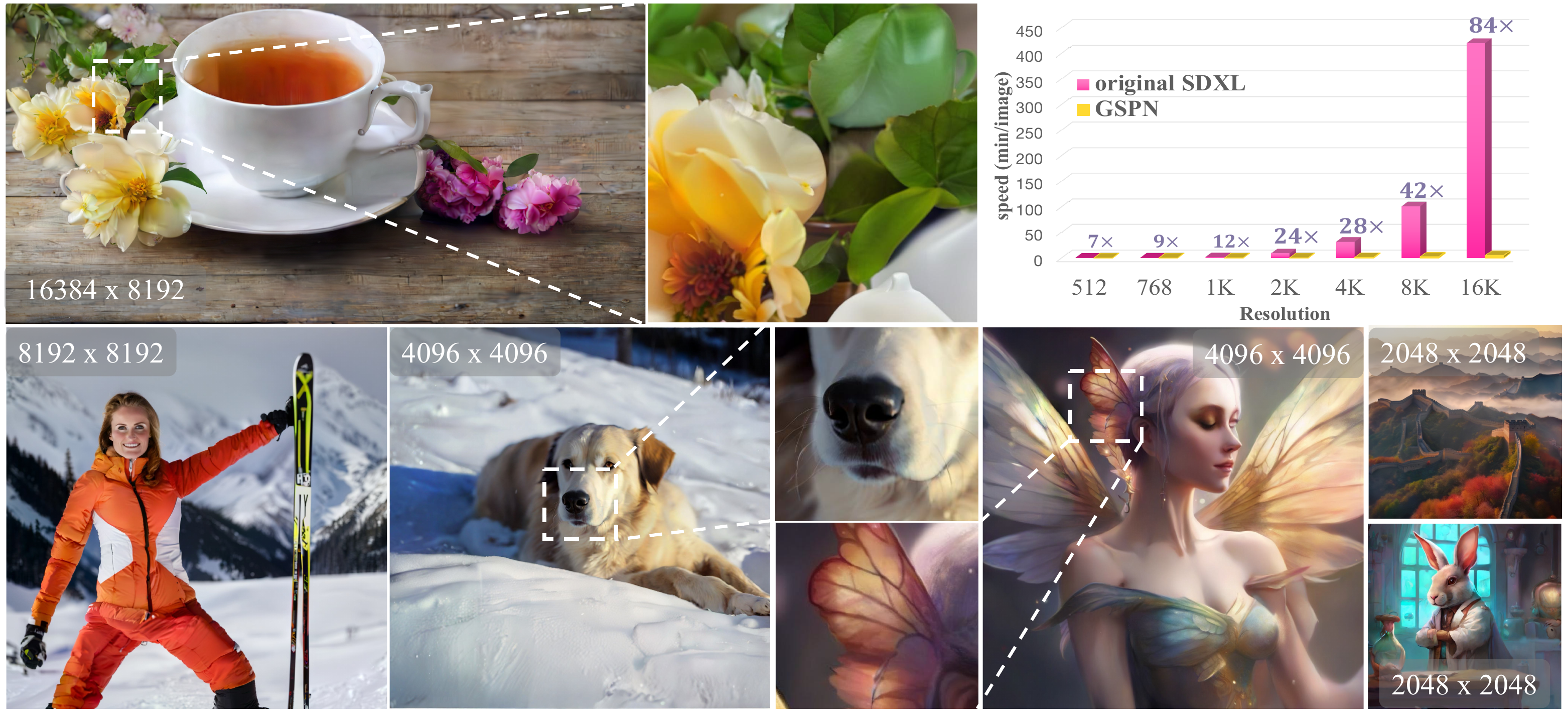}
    \caption{\textbf{Qualitative text-to-image results generated from our SD-XL-GSPN model.} We enable generation up to 16K resolution on a single A100 GPU while reducing inference time by up to $84\times$ on the SD-XL model.}
    \label{fig:vis}
\end{figure*}

\noindent\textbf{Evaluation Metrics.} For classification, we report top-1 accuracy, parameter count and computational complexity (Gflops) for comparison. For class-conditional generation, we measure scaling performance with Frechet Inception Distance (FID-50K)~\cite{heusel2017gans} using 250 DDPM sampling steps, with measurements conducted through ADM's TensorFlow evaluation protocol~\cite{dhariwal2021diffusion} to ensure accurate comparisons~\cite{parmar2022aliased}. Supplementary metrics include Inception Score~\cite{salimans2016improved}, sFID~\cite{nash2021generating} and Precision/Recall~\cite{kynkaanniemi2019improved} measurements. For text-to-image, we also consider the text similarity in the CLIP-ViT-G feature space to measure text-image alignment.

\subsection{Image Classification}
\label{sec:exp_cls}
In~\Cref{tab:cls-imagenet}, we present a comparative analysis of ImageNet-1K classification performance across three architectural paradigms: ConvNet-based~\cite{liu2022convnet,yu2024mambaout}, Transformer-based~\cite{touvron2021training,liu2021swin,dai2021coatnet,dong2022cswin,li2022next,li2022uniformer}, and sequential-based~\cite{liu2024vmamba,huang2024localmamba,liu2024vmamba,li2024mamba,duan2024visionrwkv,alkin2024visionlstm} models with different sizes. 

Our GSPN outperforms existing sequential models with comparable parameter counts. Notably, GSPN-T achieves approximately 1\% higher absolute accuracy than Vim-S (80.5\%) and VMamba-T (82.2\%). With lower computational complexity (5.3 GFLOPs \vs 5.7 GFLOPs), GSPN-T surpasses LocalVMamba-T by 0.3\% points while maintaining competitive performance against SoTA ConvNet-based and Transformer-based architectures. While GSPN-S performs slightly below MambaOut-Small, we observe that MambaOut struggles to scale effectively as parameters double (compared to MambaOut-Base), which limits its practical scalability. The consistent performance improvements across model scales, from tiny to base configurations, demonstrate GSPN's scalability characteristics comparable to ViT, establishing it as a viable alternative to traditional vision transformers.

\subsection{Class-conditional Generation}
\label{sec:exp_gen}

\input{tables/gen}
We compare GSPN with several competitive diffusion transformer methods~\cite{bao2023all,Peebles2023DiT,chen2023pixart,ma2024sit}. For a fair comparison, we evaluate models with models of comparable parameter budgets trained over 400K iterations without classifier-free guidance~\cite{ho2022classifier} in~\Cref{tab:gen}. Our GSPN-XL/2 establishes new SoTA performance, while GSPN-L/2 achieves superior FID scores on ImageNet 256×256 class-conditional generation with merely 65.6\% of the parameters compared to prior models. Extensive scaling studies validate GSPN's efficiency, with GSPN-B/2 achieving competitive performance at 20.3\% of DiT-XL/2's parameter count when converging, highlighting GSPN's promising efficiency and scalability.

\subsection{Text-to-image Generation}
\label{sec:exp_t2i}

In~\Cref{tab:t2i}, we compare GSPN with Mamba~\cite{gu2023mamba}, Mamba2~\cite{dao2024transformers}, and Linfusion~\cite{liu2024linfusion} using SD-v1.5~\cite{rombach2022high} as the baseline. On the COCO benchmark at $1024\times1024$ resolution (unseen during training), GSPN achieves superior FID and CLIP scores. Unlike Mamba and Linfusion, which require extra normalization for unseen resolutions, GSPN adapts to arbitrary resolutions due to its normalized weights $W_{ij}$ satisfying the Stability-Context Condition (see Theorem \ref{the:1}). This property ensures stable and effective long-range propagation without extra normalization during training or inference, validating GSPN’s ability to capture broader spatial dependencies efficiently. Notably, GSPN matches the performance of Linfusion \cite{liu2024linfusion} without using any pretrained weights and within the same number of training epochs.
\input{tables/t2i}

\subsection{Ablation Study}
We conduct ablation studies on GSPN-T for image classification and GSPN-B for class-conditional generation (\Cref{fig:ablation}) to identify the optimal architecture design. Parameter counts are kept consistent across configurations for fair comparison.

\noindent\textbf{Hybrid Design.} Starting with an all-local GSPN baseline achieving 81.2\% Top-1 accuracy and 27.2 FID, incorporating global GSPN in the network's top layers brings consistent improvements across tasks.

\noindent\textbf{Micro Design.} Replacing SPN's~\cite{liu2017learning} absolute normalization and max-pooling merging with sigmoid and learnable merging improves Top-1 accuracy to 82.3\%. Removing PE and GLU while adding layers to maintain parameter count yields another 2\% gain.

\noindent\textbf{Macro Design.} Increasing hidden dimensions by $1.5\times$ and adding GRN layers contributes an additional 1.0\% in ImageNet accuracy. For generation, we achieve further improvements of 2.8\%, 1.6\%, and 1.8\% by removing adaLN-zero~\cite{Peebles2023DiT}, adding skip connections between shallow and deep layers~\cite{rombach2022high}, and replacing MLP with FFN containing depthwise convolution and GLU.

\subsection{Qualitative Results}
We present qualitative examples to illustrate the effectiveness of GSPN on ultrahigh-resolution generation in~\Cref{fig:vis}. To avoid distortion and duplication~\cite{huang2024fouriscale,he2024scalecrafter}, we build GSPN upon DemoFusion~\cite{du2024demofusion}, a pipeline dedicated to high-resolution generation from low resolutions. We also eliminate the need for patch-wise inference~\cite{bar2023multidiffusion,du2024demofusion,lin2024cutdiffusion,lin2024accdiffusion,haji2024elasticdiffusion} as GSPN is more efficient than self-attention. Inspired by~\cite{meng2021sdedit,liu2024linfusion}, we skip 60\% of initial denoising steps in the high-resolution stage, leveraging low-resolution structural information for enhanced efficiency. This optimization enables generation up to 16K resolution on a single A100 GPU, achieving $\sim84\times$ speedup at 16K$\times$8K resolution. Furthermore, we can enjoy distributed parallel inference by directly applying DistriFusion~\cite{li2024distrifusion} to GSPN, contrasting with traditional attention-based approaches that require extensive token transmission.

\subsection{Efficiency and Limitation}
In Figure \ref{fig:teaser}, we compare the inference speed of various attention layers, setting the feature dimension to 1 for memory constraints. GSPN shows comparable speed to other methods for input resolutions below 2K, where speedup is minimal. Beyond 2K, GSPN, particularly the local module, scales linearly with image size, providing significant advantages over transformer-based and linear attention.

In practice, however, our customized CUDA kernel performs slower than theoretical efficiency as the channel and batch size increase, due to inefficient value access and lack of shared memory usage. A detailed discussion on the parallel time complexity and limitations of our CUDA implementation is provided in the appendix.

%% file: tables/gen.tex
\begin{table}[!h]
  \caption{\textbf{Comparing GSPNs against competitive diffusion architectures on ImageNet 256$\times$256 generation.} All models are aligned to DiT's 400K-iteration setting without classifier-free guidance for fairness.
  }
  \label{tab:gen}
  \centering
  \footnotesize
\scalebox{0.78}{
\begin{tabular}{lcccccc}
  \toprule
  \multicolumn{7}{l}{\textbf{Class-Conditional ImageNet 256$\times$256}} \\
  \toprule
  Model & \# Params (M) & FID$\downarrow$   & sFID$\downarrow$  & IS$\uparrow$     & Precision$\uparrow$ & Recall$\uparrow$ \\
  \midrule
  DiT-XL/2~\cite{Peebles2023DiT} & 675 & 20.05 & 6.87 & 64.74 & 0.621 & 0.609 \\
  U-ViT-H/2~\cite{bao2023all} & 641 & 21.71 & 7.24 & 62.76 & 0.608 & 0.584 \\
  PixArt-$\alpha$-XL/2~\cite{chen2023pixart} & 650 & 24.81 & \underline{6.38} & 51.76 & 0.603 & 0.615 \\
  SiT-XL/2~\cite{ma2024sit} & 675 & 18.04 & \textbf{5.17} & 73.90 & 0.630 & \textbf{0.640} \\
  \midrule
  \rowcolor{Green}
  \textbf{GSPN-B/2 (Ours)} & 137 & 28.70 & 6.87 & 50.12 & 0.585 & 0.609 \\
  \rowcolor{Green}
  \textbf{GSPN-L/2 (Ours)} & 443 & \underline{17.25} & 8.78 & \underline{77.37} & \underline{0.657} & 0.417 \\
  \rowcolor{Green}
  \textbf{GSPN-XL/2 (Ours)} & 690 & \textbf{15.26} & 6.51 & \textbf{85.99} & \textbf{0.670} & \underline{0.624} \\
  \bottomrule
  \end{tabular}
}
\end{table}

%% file: tables/t2i.tex
\begin{table}[!h]
\centering
\caption{\textbf{Cross-resolution generation on the COCO benchmark under $1024\times1024$ resolution.} Lower FID ($\downarrow$) and higher CLIP-T ($\uparrow$) stand for better image quality and text-image alignment.} %
\label{tab:t2i}
\centering
\footnotesize
\scalebox{1.0}{
\begin{tabular}{l|cc}
\toprule
Model & FID($\downarrow$)    & CLIP-T($\uparrow$) \\ 
\toprule
SD-v1.5 (baseline)  & 32.71  & 0.290 \\ 
Mamba~\cite{gu2023mamba} (w/ norm) & 50.30  & 0.263 \\ 
Mamba2~\cite{dao2024transformers} (w/ norm) & 37.02  & 0.273 \\ 
Linfusion~\cite{liu2024linfusion} (w/ norm) & 36.33  & 0.285 \\ 
\midrule
\rowcolor{Green}
\textbf{SD-v1.5-GSPN w/o init (Ours)} & 36.89 & 0.278 \\
\rowcolor{Green}
\textbf{SD-v1.5-GSPN (Ours)} & \textbf{30.86}  & \textbf{0.307} \\
\bottomrule
\end{tabular}
}
\end{table}

%% file: sec/5_con.tex
\section{Conclusion}
We introduce the Generalized Spatial Propagation Network (GSPN), a novel attention mechanism for parallel sequence modeling in vision tasks. With the Stability-Context Condition ensuring stable, context-aware propagation, GSPN reduces sequence length to \( \sqrt{N} \) while maintaining efficiency. Featuring learnable, input-dependent weights without positional embeddings, GSPN overcomes the limitations of transformers, linear attention, and Mamba, which treat images as 1D sequences. GSPN achieves state-of-the-art results and significant speedups, including over 84× faster generation of 16K images with SD-XL, demonstrating its efficiency and potential for vision tasks.

%% file: sec/X_supp.tex
\setcounter{page}{1}
\maketitlesupplementary

\section{Stable-Context Condition}
In this section, we present comprehensive mathematical proofs for the Stable-Context Condition introduced in Section \ref{sec:condition} in the main paper. We revisit 2D linear propagation:
\begin{equation}
    h_i^c = w_i^c h_{i-1}^c + \lambda_i^c \odot x_i^c, \quad i \in [1, n-1], \; c \in [0, C-1]
    \label{eq:2dsp}
\end{equation}
where $i$ denotes the $i_{th}$ row and $w_i\in \mathbb{R}^{n\times n}$ denotes the tri-diagonal matrix associating it to the $(i-1)_{tj}$ row. While $\lambda_{i}$ denotes a vector to weigh $x_i$ with element-wise product, it can also be formulated as a diagonal matrix with $\lambda_{i}$'s original values being on the main diagonal. To expand Eq. \eqref{eq:2dsp}, we denote $H\in \{h_i\}, i\in [0, N-1], N=n^2$ as the latent space where the propagation is carried out:
\begin{small}
	\begin{equation}
	H_v =
	\left[
	\begin{matrix}
	I     & 0    & \cdots & \cdots & 0    \\
	w_2    & \lambda_2  & 0 & \cdots & \cdots   \\
	w_3w_2     & w_3\lambda_2    & \lambda_3  &  0  & \cdots 		\\
	\vdots &   \vdots   & \vdots & \ddots & \vdots    \\
	\vdots &   \vdots   & \cdots & \cdots    & \lambda_{N-1}    \\
	\end{matrix}
	\right]X_v	\\
	=GX_v,
	\label{eq:global}
	\end{equation}
\end{small}
Here, $G$ is a lower triangular, $N\times N$ transformation matrix relating $X$ and $H$, and $X_v$ and $H_v$ are vectorized sequences concatenating all the rows, i.e., $\left[x_0,x_1,...,x_{n-1}\right]$ and $\left[h_0,h_1,...,h_{n-1}\right]$, with length of $N$. We denote each $n\times n$ block as one sub-matrix, on setting $\lambda_0 = I$, the $i^{th}$ constituent $n \times n$ sub-matrix of $G_{ij}$ is:
\begin{equation}
    G_{ij}=\left\{
    \begin{aligned}
    \prod_{\tau=j+1}^{i}&w_\tau\lambda_j, &j\in[0,i-1] \\
    &\lambda_j,\qquad &i=j 
    \end{aligned}
    \right.
    \label{eq:G}
\end{equation}

Thus, \( W_{ij} = \prod_{\tau=j+1}^{i} w_\tau \) is crucial for maintaining dense connections between the elements in \( X \) and \( H \). As noted in the main paper, \( W_{ij} = \prod_{\tau=j+1}^{i} w_\tau \) must avoid collapsing to a matrix with a small norm to enable effective long-context propagation. This ensures that \( h_i \) can substantially contribute to \( h_j \) even over extended intervals, preserving the influence of distant elements. To achieve this, we design $W_{ij}$ to satisfy $\sum_{j=0}^{n-1}W_{ij}=1$, ensuring that each element in $h_i$ is a weighted average of all the elements of $x'_{j}=\lambda_jx_j$. This design guarantees that the information in the $j^{th}$ column is not diminished when propagated to the $i^{th}$ column.

\begin{figure*}[t]
    \centering
    \includegraphics[width=\linewidth]{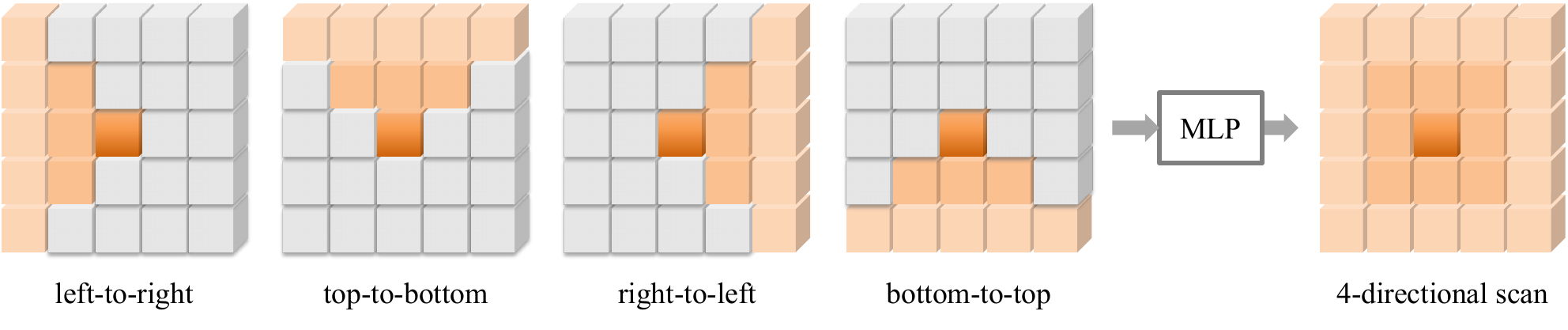}
    \caption{GSPN guarantees \textbf{Dense Pairwise Connections} via 3-way connection and 4-directional scanning, as introduced in Sec \ref{sec:condition} and detailed in Sec. \ref{sec:dense}. The scanning of each direction corresponds to a lower triangular affinity matrix. The finally full matrix is obtained through a learnable linear aggregation, denoted as ``MLP'' in the figure.}
    \label{fig:3way4dir}
\end{figure*}
\subsection{Long-context Condition}
\begin{theorem}\label{the:01}
    If all the tridiagonal matrices \( w_{\tau} \) are row stochastic, then \(\sum_{j=0}^{n-1} W_{ij} = 1\) is satisfied.
\end{theorem}
\paragraph{Definition of Row Stochastic Matrix.}
A matrix \( T \) is row stochastic if:
\begin{itemize}
    \item All elements are non-negative: \( T_{ij} \geq 0 \) for all \( i, j \).
    \item The sum of the elements in each row is 1: \( \sum_{j} T_{ij} = 1 \) for all \( i \).
\end{itemize}

\paragraph{Proof.}
Let \( A \) and \( B \) be two \( n \times n \) row stochastic matrices. We need to show that their product \( C = AB \) is also row stochastic.

\noindent\textbf{Non-negativity}: Since \( A \) and \( B \) are both nonnegative, all elements of their product \( C \) are also nonnegative:
\[
C_{ik} = \sum_{j=1}^{n} A_{ij} B_{jk} \geq 0
\]
because each \( A_{ij} \geq 0 \) and each \( B_{jk} \geq 0 \).

\noindent\textbf{Row Sum}: We need to show that the sum of the elements in each row of \( C \) is 1. Consider the \( i \)-th row sum of \( C \):
\[
\sum_{k=1}^{n} C_{ik} = \sum_{k=1}^{n} \sum_{j=1}^{n} A_{ij} B_{jk}
\]
By changing the order of summation:
\[
\sum_{k=1}^{n} C_{ik} = \sum_{j=1}^{n} A_{ij} \left( \sum_{k=1}^{n} B_{jk} \right)
\]
Since \( B \) is row stochastic:
\[
\sum_{k=1}^{n} B_{jk} = 1
\]
Therefore:
\[
\sum_{k=1}^{n} C_{ik} = \sum_{j=1}^{n} A_{ij} \cdot 1 = \sum_{j=1}^{n} A_{ij}
\]
And since \( A \) is row stochastic:
\[
\sum_{j=1}^{n} A_{ij} = 1
\]
Hence:
\[
\sum_{k=1}^{n} C_{ik} = 1
\]
This shows that \( C = AB \) is row stochastic.

\paragraph{Induction for Multiple Matrices.} To extend this result to the product of several row stochastic matrices, we proceed by induction.

\noindent\textbf{Base Case:} The product of two row stochastic matrices is row stochastic, as shown above.

\noindent\textbf{Inductive Step:} Assume that the product of \( m \) row stochastic matrices \( w_1 w_2 \cdots w_m \) is row stochastic. We need to show that \( w_1 w_2 \cdots w_m w_{m+1} \) is also row stochastic.

By the induction hypothesis, \( W = w_1 w_2 \cdots w_m \) is row stochastic. Since \( w_{m+1} \) is also row stochastic, the product \( W w_{m+1} \) follows the properties proven above for the product of two row stochastic matrices.

\noindent\textbf{Conclusion:} By induction, the product of any finite number of row stochastic matrices is row stochastic. Since tridiagonal matrices form a subset of such matrices, the proof applies to them as well. This ensures that \( \sum_{j=0}^{n-1} W_{ij} = 1 \) holds for all \( i \).

\subsection{Stability Condition}

\begin{theorem}\label{the:02}
    The stability of Eq.~\eqref{eq:2dsp} is ensured when all matrices $w_{\tau}$ are row stochastic.
\end{theorem}

By ignoring $c$ for simplicity, we re-write Eq. \eqref{eq:2dsp}, where each $h_{k,i}$ is computed as the following. We will use $p_{k,i}$ in the proof of Theorem \ref{the:02}.
\begin{equation}
h_{k,i} = \lambda_{k,i} x_{k,i} + \sum_{k\in\mathbb{N}}p_{k,i}h_{k,i-1}
\label{eq:2rnn-spatial}
\end{equation}

\paragraph{Proof.}
The stability of linear propagation refers to preventing responses or errors from growing unbounded and ensuring that gradients do not vanish during backpropagation, as described in \cite{liu2017learning}. Specifically, for a stable model, the norm of the temporal Jacobian \( \frac{\partial h_i}{\partial h_{i-1}} \) should be less than or equal to 1. In our case, this requirement can be met by ensuring that the norm of each transformation matrix \( w_i \) satisfies:
\[
\left\| \frac{\partial h_i}{\partial h_{i-1}} \right\| = \| w_i \| \leq \sigma_{\text{max}} \leq 1,
\]
where \( \sigma_{\text{max}} \) denotes the largest singular value of \( w_i \). This condition ensures stability.

By Gershgorin's Circle Theorem, every eigenvalue \( \sigma \) of a square matrix \( w_i \) satisfies:
\[
|\sigma - p_{i,i}| \leq \sum_{k=1, k \neq i}^{n} |p_{k,i}|, \quad i \in [1, n],
\]
which implies:
\[
\sigma_{\text{max}} \leq |\sigma - p_{i,i}| + |p_{i,i}| \leq \sum_{k=1}^{n} |p_{k,i}|.
\]

If \( w_i \) is row stochastic, then:
\[
\sum_{k=1}^{n} |p_{k,i}| = 1 \quad \text{for each row}.
\]

Thus:
\[
\sigma_{\text{max}} \leq 1,
\]
satisfying the stability condition. Making \( w_{\tau} \) row stochastic ensures that the norm constraint \( \| w_i \| \leq 1 \) holds, which provides a sufficient condition for model stability, as presented in \cite{liu2017learning}.

\begin{figure*}[!t]
    \centering
    \includegraphics[width=\linewidth]{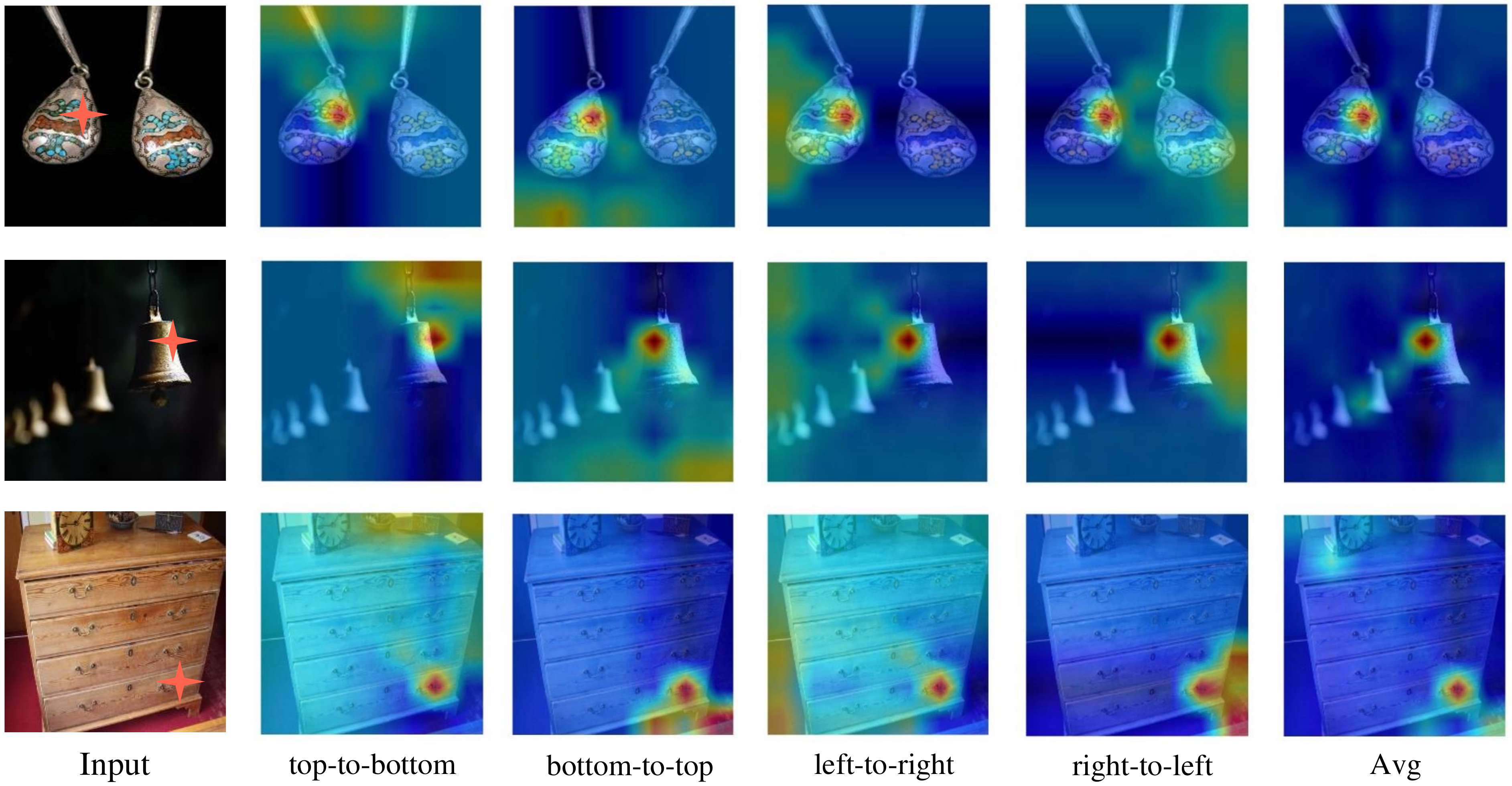}
    \caption{Illustration of heatmaps for the query patch (marked with an orange star) along different directions and the averaged results.}
    \label{fig:heatmap}
\end{figure*}

\subsection{Guaranteeing Dense Pairwise Connections} \label{sec:dense}
In this section, we provide an intuitive rationale for selecting the 3-way connection, represented by a tri-diagonal matrix, as the minimal structure needed for dense matrix production through multiplication. To propagate non-zero entries throughout a matrix, sufficient connectivity is essential. Diagonal matrices, which have non-zero elements only on the main diagonal, fail to propagate non-zero values to off-diagonal positions, resulting in products that remain sparse. Similarly, matrices with non-zero elements limited to the main diagonal and one adjacent diagonal (e.g., upper or lower bi-diagonal matrices) restrict the spread of non-zero entries, maintaining a banded structure even after multiplication.

In contrast, tri-diagonal matrices, with non-zero elements on the main diagonal as well as the adjacent upper and lower diagonals, enable significant propagation of non-zero values during multiplication. This 3-way connection facilitates the spread of non-zero entries beyond the initial three bands. When two tri-diagonal matrices are multiplied, their non-zero entries extend further, and repeated multiplications lead to more comprehensive filling of the resulting matrix. Consequently, the tri-diagonal structure represents the minimal configuration with sufficient off-diagonal connections to eventually produce a dense matrix. This makes tri-diagonal matrices the simplest form capable of ensuring dense propagation in matrix products.

In Figure \ref{fig:3way4dir}, we show the 3-way connection in 4 different directions. The scanning of each direction corresponds to a lower triangular affinity matrix, i.e., $G$ in Eq. \eqref{eq:global}. A full dense matrix is obtained through learnable aggregation via a linear layer, as described in Sec. \ref{sec:principle}, guaranteeing dense pairwise connections with sub-linear complexity.

\section{Attention Heatmaps of GSPN}
The heatmap in~\Cref{fig:heatmap} presents a comprehensive analysis of our GSPN across four distinct directional scans, revealing a pronounced anisotropic behavior. Each directional heatmap represents a unique lower triangular affinity matrix. An additional aggregated heatmap provides a holistic view, synthesizing the directional insights into a unified representation that captures long-context and dense pairwise connections through a 3-way connection mechanism.

\section{CUDA Implementation}
We implement a highly-parallel computation for GSPN on CUDA-enabled GPUs, consisting of forward and backward passes. The detailed process is shown in~\Cref{alg:gspn}.

\noindent\textbf{Forward Pass.} For each position $(n,c,h,w)$ in the 4D tensor, we compute three directional connections (diagonal-up, horizontal, diagonal-down) using gates $G1$, $G2$, and $G3$. These operations are equivalent to matrix multiplication between $h$ and $w$ dimensions (Figure 2). The computation is divided into $g$ groups, where $g=1$ indicates global GSPN, and $g>1$ indicates local GSPN. The final hidden state $H$ combines input transformation $x_{hype} = B × X$ with directional connections $h_{hype}$.

\noindent\textbf{Backward Pass.} Gradients are computed for all inputs $(X, B, G1, G2, G3)$ by reverse propagation. For each position, gradients flow from future timesteps into $h_{diff}$. Input gradients $X_{diff}$ are computed via $B$ values, while gate gradients $(G1_{diff}, G2_{diff}, G3_{diff})$ use error terms and previous hidden states.

Both passes leverage CUDA's parallel processing by distributing computations across threads. As the inner loop operates in parallel, GSPN's complexity is determined by the outer loop, resulting in $\mathcal{O}(\sqrt{N})$ complexity.
\begin{algorithm}
\caption{Forward Pass CUDA Implementation}
\begin{algorithmic}[1]
\Require Input tensors X, B, G1, G2, G3
\Require width, kNItems
\Ensure Output tensor H
\State g = $\lceil$width / kNItems$\rceil$
\State count = g × height × channels × num

\For{t = 0 to kNItems - 1}
    \ParFor{index = 0 to count - 1}
        \State Calculate n, c, h, k from index
        \State w = k × nitems + t
        
        \State x\_data = X[n,c,h,w]
        \State b\_data = B[n,c,h,w]
        
        \State h1  = G1[n,c,h,w,h-1,w-1] × H[n,c,h-1,w-1]
        \State h2  = G2[n,c,h,w,h,w-1] × H[n,c,h,w-1]
        \State h3  = G3[n,c,h,w,h+1,w-1] × H[n,c,h+1,w-1]
        
        \State h\_hype = h1  + h2  + h3 
        \State x\_hype = b\_data × x\_data
        \State H[n,c,h,w] = x\_hype + h\_hype
    \EndParFor
\EndFor
\end{algorithmic}
\label{alg:gspn}
\end{algorithm}

\begin{algorithm}
\caption{Backward Pass CUDA Implementation}
\begin{algorithmic}[1]
\Require Input tensors X, B, G1, G2, G3, H, H\_diff
\Require width, kNItems
\Ensure Output tensors X\_diff, B\_diff, G1\_diff, G2\_diff, G3\_diff, H\_diff
\State g = $\lceil$width / kNItems$\rceil$
\State count = g × height × channels × num

\For{t = 0 to kNItems - 1}
    \ParFor{index = 0 to count - 1}
        \State Calculate n, c, h, k from index
        \State w = width - 1 - k × nitems - t
        
        \State h\_diff = H\_diff[n,c,h,w]
        \State Update h\_diff with future timestep contributions
        \State H\_diff[n,c,h,w] = h\_diff
        
        \State X\_diff[n,c,h,w] = B[n,c,h,w] × h\_diff
        \State B\_diff[n,c,h,w] = X[n,c,h,w] × h\_diff
        
        \State G1\_diff[n,c,h,w,h-1,w-1] = h\_diff × H[n,c,h-1,w-1]
        \State G2\_diff[n,c,h,w,h,w-1] = h\_diff × H[n,c,h,w-1]
        \State G3\_diff[n,c,h,w,h+1,w-1] = h\_diff × H[n,c,h+1,w-1]
    \EndParFor
\EndFor
\end{algorithmic}
\end{algorithm}

\section{Limitation}

\begin{table*}[t!]
\centering
\small
\begin{tabular}{l|c|c|c|c}
\toprule
\textbf{Method} & \textbf{Total Seq.} & \textbf{Ideal Parallel TC} & \textbf{Practical Parallel TC $T_p$} & \textbf{MC} \\ 
\midrule
\textbf{Softmax Attention} & $O(N^2 \cdot d)$ & $O(1)$ & $O\left(\frac{N^2 \cdot d}{P}\right)$ & $O(N^2)$ \\ 
\textbf{Linear Attention} & $O(N \cdot d)$ & $O(1)$ & $O\left(\frac{N \cdot d}{P}\right)$ & $O(N \cdot d)$ \\ 
\textbf{Mamba} & $O(N \cdot d)$ & $O(N)$ & $O(N \cdot d)$ & $O(d)$ \\ \midrule
\textbf{GSPN-global} & $O(4N \cdot d)$ & $O(4\sqrt{N})$ & $O\left(\frac{4N \cdot d}{P}\right)$ & $O(4\sqrt{N} \cdot d)$ \\ 
\textbf{GSPN-local} & $O(\frac{4N \cdot d}{g})$ & $O(\frac{4\sqrt{N}}{g})$ & $O\left(\frac{4N \cdot d}{g\cdot P}\right)$ & $O(\frac{4\sqrt{N} \cdot d}{g})$ \\ \bottomrule
\end{tabular} \vspace{2mm}
\caption{Complexity analysis for different attention mechanisms and propagation methods. ``TC'' denotes time complexity, and ``MC'' denotes memory complexity.}
\label{tab:complexity}
\end{table*}
The main limitation of the current GSPN framework lies in the optimization of memory access in our customized CUDA kernel implementation. The hidden vector $H$, frequently accessed by multiple threads, is stored in global memory without utilizing shared memory, leading to inefficient memory access patterns, increased latency, and higher contention for global memory bandwidth, particularly at high resolutions. Additionally, the lack of coalesced memory access and reliance on redundant index computations further degrade performance, especially as channel and batch sizes increase.

While efficiency analysis in~\Cref{fig:teaser} highlights GSPN's scalability, demonstrating significant advantages over transformer-based and linear attention approaches for image resolutions beyond 2K, it reveals limited efficiency gains for low-resolution inputs. This discrepancy stems from implementation inefficiencies in our CUDA kernel, such as suboptimal value access and the absence of shared memory optimization. Addressing these bottlenecks is critical for fully leveraging GSPN’s theoretical efficiency across a wider range of input sizes.

\input{tables/gen_stat}
\input{tables/cls_stat}
\section{Parallel Time and Memory Complexity}
In this section, we provide a detailed analysis of the computational complexity for various attention mechanisms and propagation methods used in handling 2D data, focusing on their time and memory efficiency.
\noindent\textbf{Softmax Attention.}
Softmax attention is the most classical attention model. It computes full pairwise interactions between all $N$ pixels in a 2D grid, resulting in a total sequential work of $O(N^2 \cdot d)$, where $d$ is the feature dimension. While its ideal parallel time complexity is $O(1)$ due to independent pairwise computations, practical parallel time complexity is $O\left(\frac{N^2 \cdot d}{P}\right)$, where $P$ represents the number of available GPU cores, and its memory complexity is $O(N^2)$. 

\noindent\textbf{Linear Attention.} Linear attention reduces complexity by computing $Q(K^T V)$, which avoids the need for full pairwise interactions seen in softmax attention. This factorization allows for a linear computational cost of $O(N \cdot d)$, where $N$ is the total number of pixels or tokens, and $d$ is the feature dimension. The total sequential work of $O(N \cdot d)$ arises because each element in $Q$ interacts with a transformed version of $V$ (i.e., $K^T V$), which can be computed in parallel across $N$. The ideal parallel time complexity is $O(1)$ because each computation for the final result can be theoretically performed independently if there are enough processing cores. The practical parallel time, $O\left(\frac{N \cdot d}{P}\right)$, depends on the available GPU cores $P$, which distribute the computation workload. Memory complexity is $O(N \cdot d)$, as only intermediate vectors for $Q$, $K$, and $V$ need to be stored during computation, making it significantly more efficient than softmax attention for large-scale data processing.

\noindent\textbf{Mamba.} Mamba performs sequential pixel-to-pixel propagation, where each pixel depends on the result of its predecessor, leading to a total sequential work complexity of $O(N \cdot d)$, where $N$ is the number of pixels and $d$ is the feature dimension. While the strict sequential dependency limits parallelism across a single sequence, practical parallel time complexity can be expressed as $O\left(\frac{N \cdot d}{P}\right)$ when $P$ processing units are utilized, particularly in scenarios where multiple sequences or batches are processed concurrently. Memory complexity remains minimal at $O(d)$, as only the current and previous pixel states need to be stored. While Mamba is memory-efficient, its sequential nature within a single sequence makes it less scalable for high-resolution 2D data compared to more parallel-friendly methods.

\noindent\textbf{GSPN.} GSPN processes data using line scan, where each pixel in a row/column depends only on its adjacent pixels from the previous row/column through a tridiagonal matrix structure. This design leads to a total sequential work complexity of $O(4N \cdot d)$, where $N$ is the total number of pixels and $d$ is the feature dimension. The ideal parallel time complexity is $O(4\sqrt{N})$, as each row can be processed in parallel, but rows are processed sequentially. The practical parallel time is $O\left(\frac{4N \cdot d}{P}\right)$. Memory complexity is $O(4\sqrt{N} \cdot d)$, as only the current and previous rows are stored during computation. GSPN's line-wise parallelism and reduced sequence length provide a balance between computational efficiency and scalability, making it suitable for handling high-resolution 2D data more effectively than fully sequential methods like Mamba.

\begin{figure*}[!ht]
    \centering
    \includegraphics[width=\linewidth]{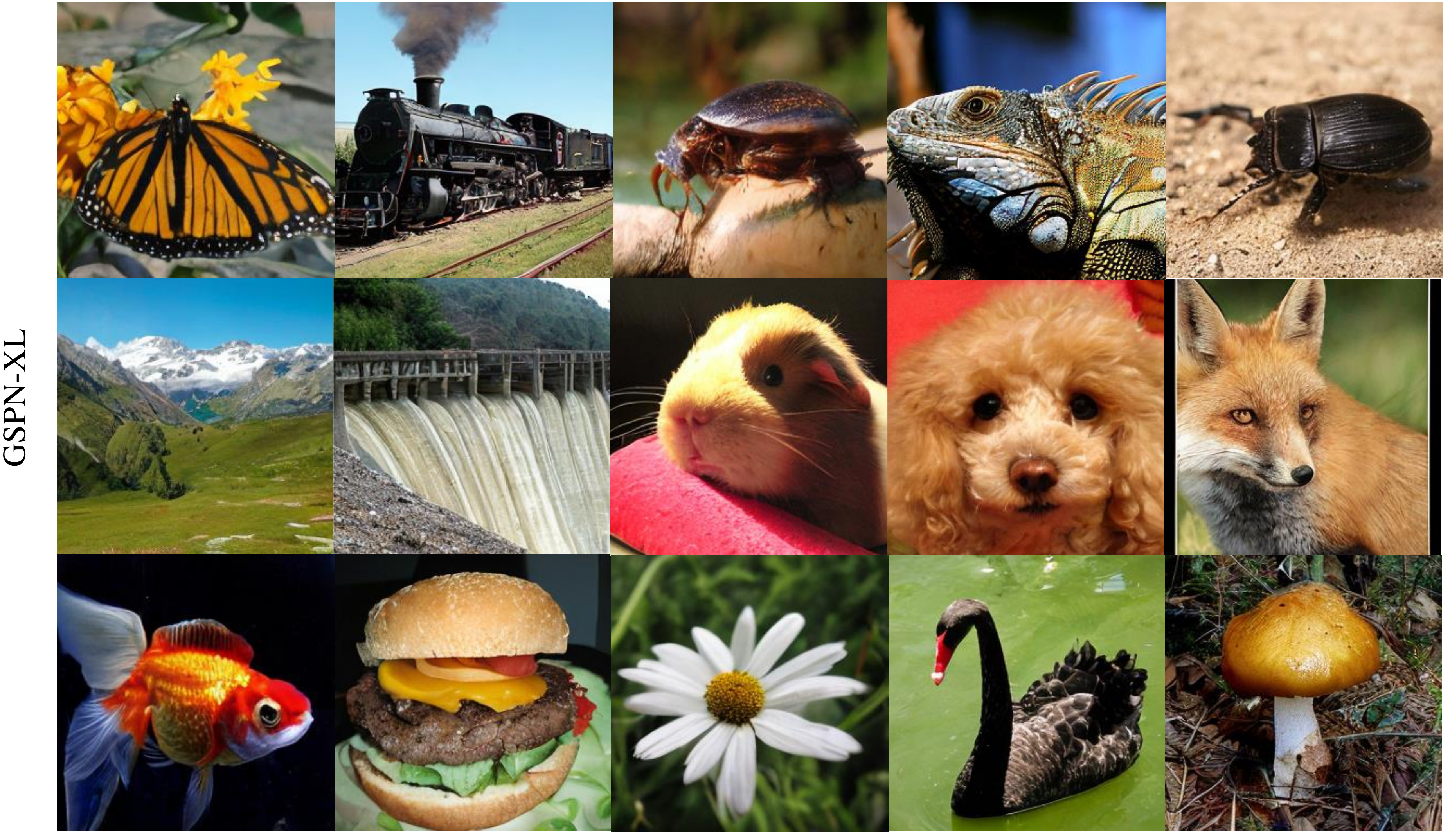}
    \caption{Qualitative results of class-conditional generation from our 256$\times$256 resolution GSPN-XL/2 models (at 400K-iteration). To ensure consistency with the quantitative results reported in the paper, classifier-free guidance was NOT applied to any of the outputs.}
    \label{fig:dit}
\end{figure*}
\section{Details of the Overall Architecture}
\noindent\textbf{Image Classification.} We implement a four-level hierarchical architecture for image classification. The initial stem module processes the $H\times W\times 3$ input through two consecutive $3\times3$ convolutions (stride 2, padding 1). The first convolution outputs half the channels of the second, with each convolution followed by LN and GELU activation.
The subsequent levels implement a progressive feature hierarchy, where levels 1-2 employ local GSPN blocks for efficient processing at higher resolutions ($H/4\times W/4$, $H/8\times W/8$), while levels 3-4 utilize global GSPN blocks for contextual integration at lower resolutions ($H/16\times W/16$, $H/32\times W/32$). Between levels, dedicated downsampling layers perform spatial reduction through $3\times3$ convolutions (stride 2, padding 1) followed by LN, where each downsampling operation halves the spatial dimensions while maintaining the hierarchical feature representation.

\noindent\textbf{Class-conditional Generation.} 
We present an elegant and versatile architecture for class-conditional generation models in Figure 3 (b). Specifically, GSPN parameterizes the noise prediction network $\epsilon_\theta(\textbf{x}_t, t, \textbf{c})$, which estimates the noise introduced into the partially denoised image, considering the timestep $t$, condition $\textbf{c}$, and noised image $\textbf{x}_t$ as inputs. The initial stage transforms the input image into flattened 2-D patches, subsequently converting them into a sequence of tokens with dimension $D$ through linear embedding. Note that there is not learnable positional embeddings in the sequence. We explore patch sizes of $p=2$ in the design space but would hold for any patch size. Beyond noised image inputs, the model incorporates additional conditional information such as noise timesteps $t$ and conditions $\textbf{c}$ like class labels or natural language. To integrate these, we append the vector embeddings of timestep $t$ and class condition as supplementary tokens in the input sequence. These tokens are added to image tokens. The hidden states from the main branch and the skip branch are concatenated and linearly projected before input to the subsequent GSPN module. The final GSPN module decodes the hidden state sequence into noise prediction and diagonal covariance prediction, preserving the original spatial input dimensions. A standard linear decoder applies the final layer normalization and linearly transforms each token, subsequently rearranging the decoded tokens to the original spatial layout.

\section{More Samples Generated from GSPN}
\begin{figure*}[!t]
    \centering
    \includegraphics[width=\linewidth]{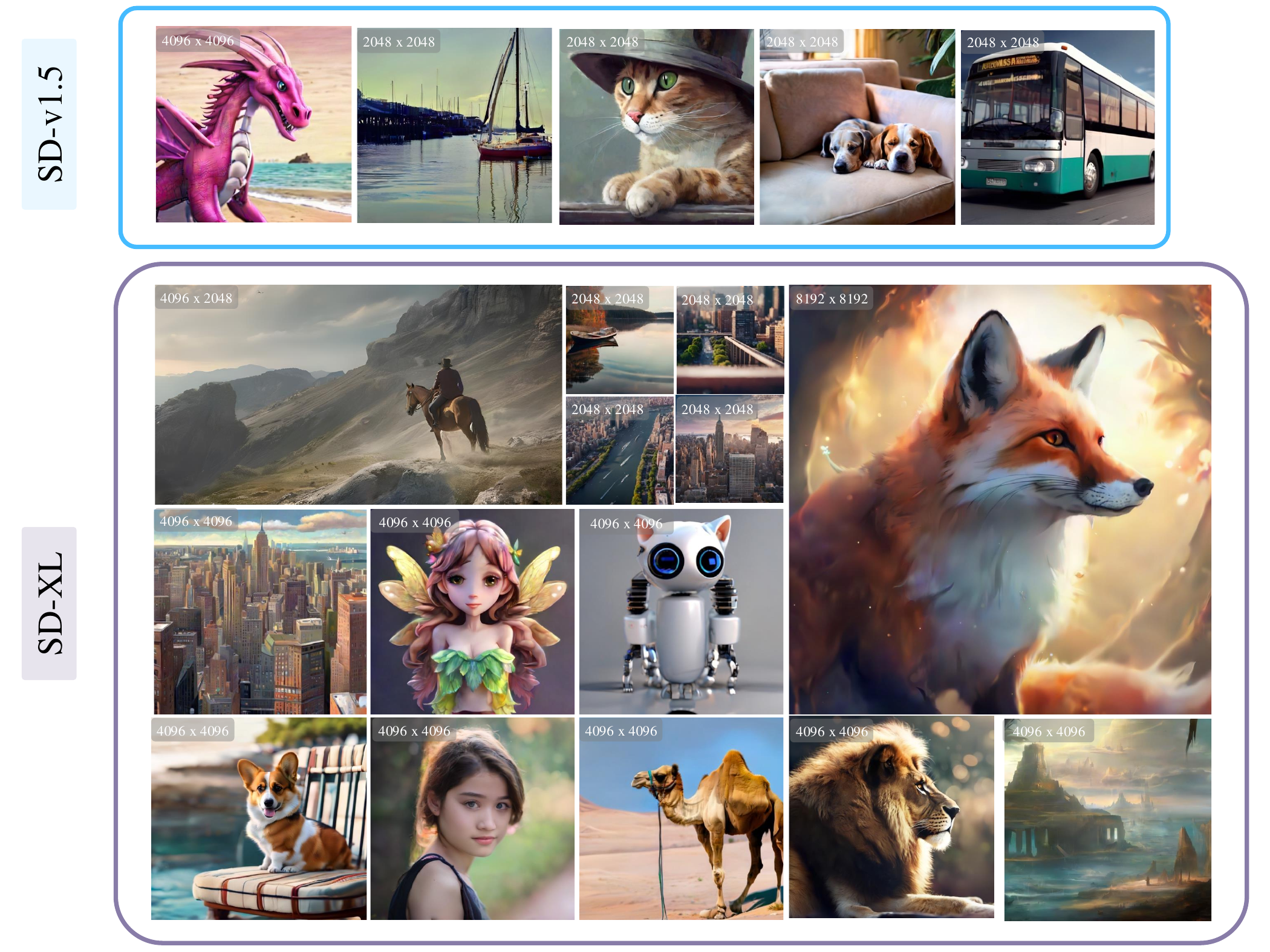}
    \caption{Examples of GSPN at various higher resolutions based upon SD-v1.5 and SDXL. GSPN enables to synthesis of images up to a resolution of 16384$\times$8192 using a single A100. Best viewed in PDF with zoom.}
    \label{fig:more_vis}
\end{figure*}

\begin{figure*}[!t]
    \centering
    \includegraphics[width=\linewidth]{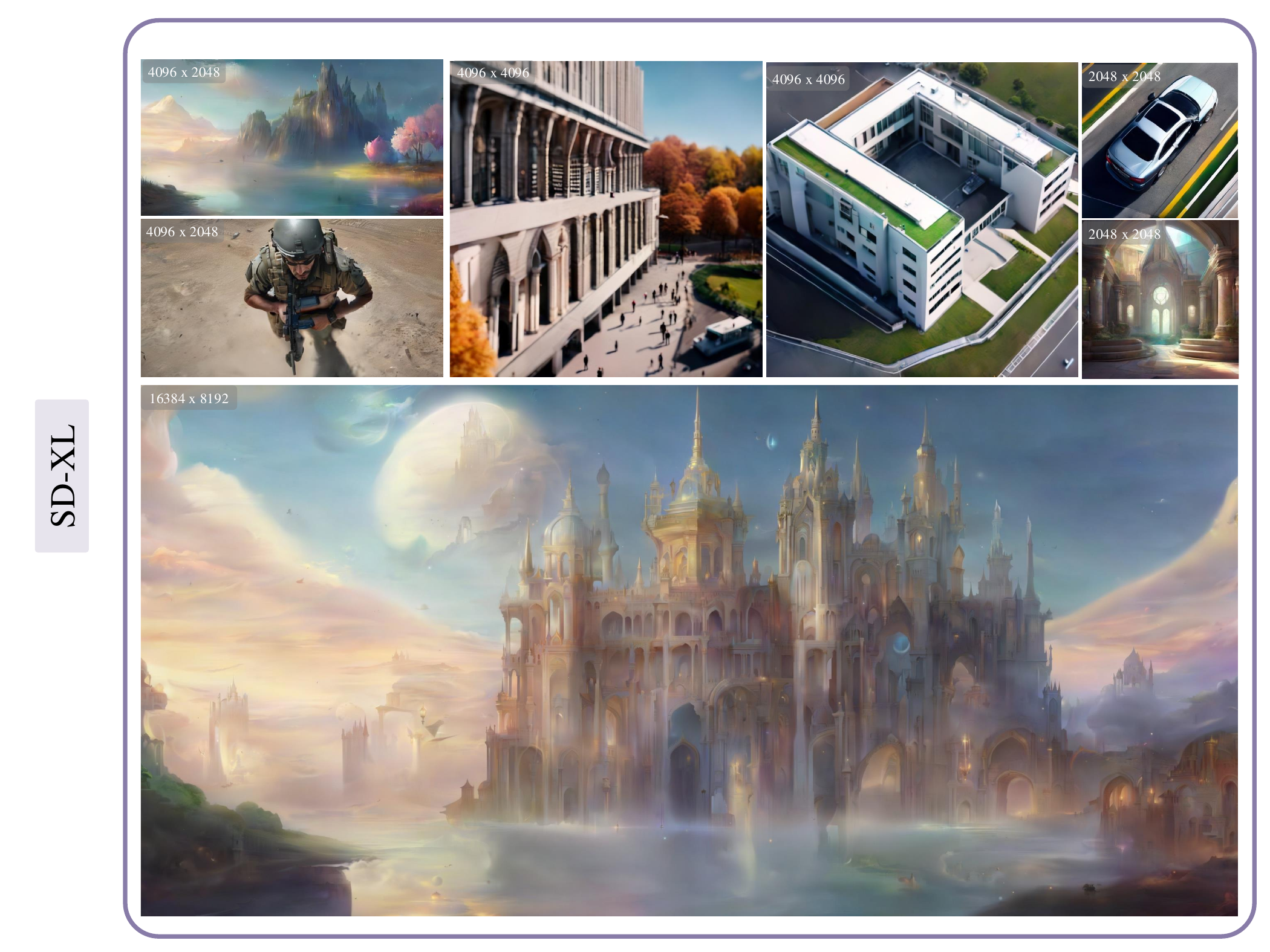}
    \caption{Examples of GSPN at various higher resolutions based upon SDXL. GSPN enables to synthesis of images up to a resolution of 16384$\times$8192 using a single A100. Best viewed in PDF with zoom.}
    \label{fig:more_vis2}
\end{figure*}

\begin{figure*}[!t]
    \centering
    \includegraphics[width=\linewidth]{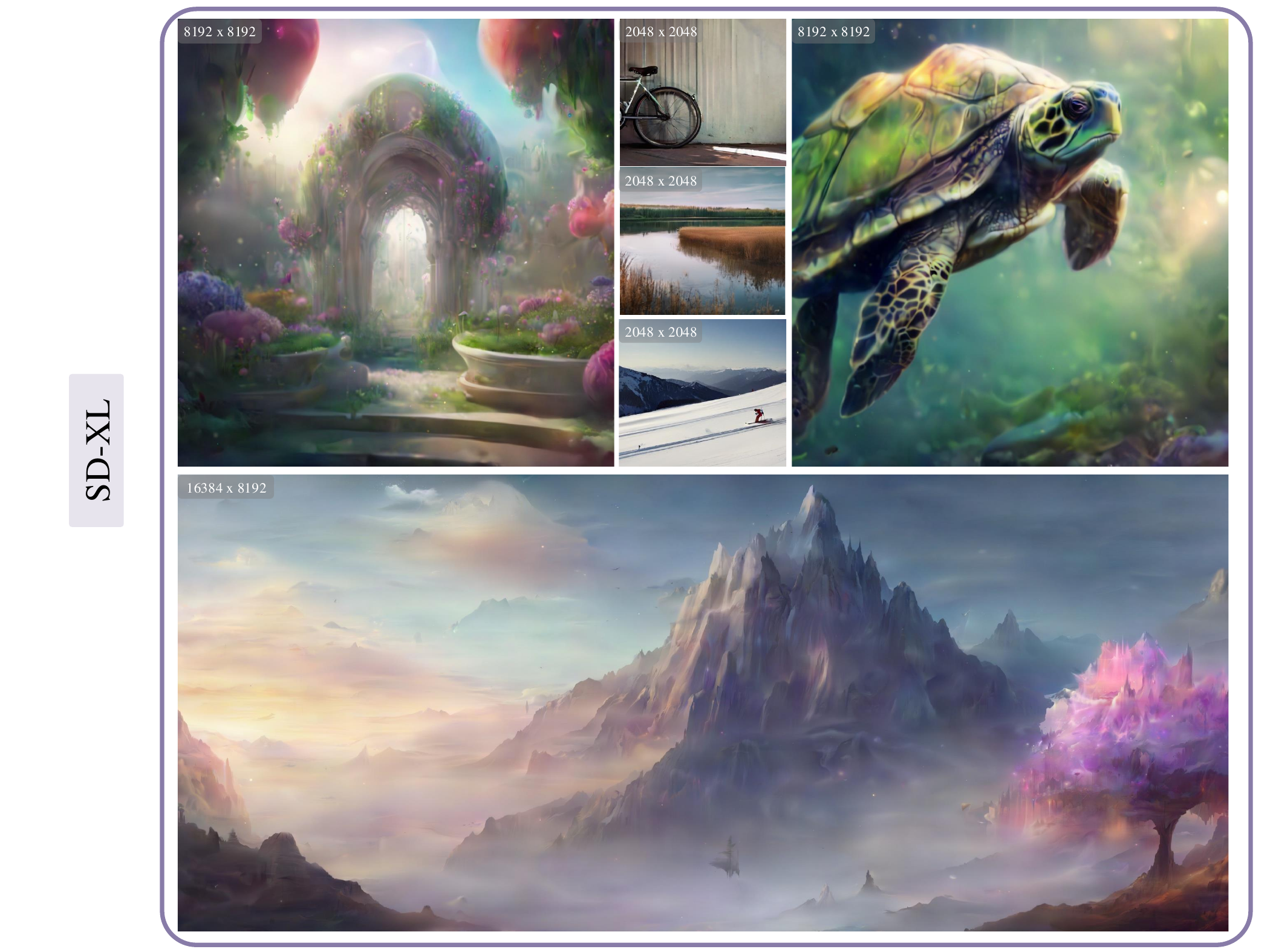}
    \caption{Examples of GSPN at various higher resolutions based upon SDXL. GSPN enables to synthesis of images up to a resolution of 16384$\times$8192 using a single A100. Best viewed in PDF with zoom.}
    \label{fig:more_vis3}
\end{figure*}

For more quantitative results on the T2I model, please zoom in to see the generated images from Figure. \ref{fig:more_vis}, \ref{fig:more_vis2}, and \ref{fig:more_vis3}.

%% file: tables/gen_stat.tex
\begin{table}[t!]
\renewcommand{\arraystretch}{1.2}
    \caption{
    \textbf{Details of GSPN models for class-conditional generation.} We follow DiT model configurations for the Base (B), Large (L) and XLarge (XL) variants. Steps/sec is measured on ImageNet 256×256 generation with patch size equal to 2 with an A100.}
    \label{tab:gen_stat}
    \centering
    \small
    \scalebox{1.0}{
    \begin{tabular}{l c c c c c}
    \toprule
    Model            & \#Layers & Hidden size &  \#Params (M)  & Steps/s \\
    \midrule 
    GSPN-B   &   30   &      900    &   137   &   2.78  \\
    GSPN-L  &    56   &      1200    &   443   &  1.15  \\
    GSPN-XL &    56  &       1500     &   690  &  0.88  \\
    \bottomrule
    \end{tabular}}
    \label{tab:models}
    \vspace{-3mm}
\end{table}

%% file: tables/cls_stat.tex
\begin{table*}[h!]
\renewcommand{\arraystretch}{1.3}
\caption{\textbf{Details of GSPN models for image classification.} \textbf{G} = global GSPN, while \textbf{L} = local GSPN.}
\label{table:transnext_config}
    \centering
    \small
    \scalebox{1.0}{
    \begin{tabular}{l c c c c c}
    \toprule
    Model            & \#Layers & Hidden size &  MLP ratio  & Mixing type \\
\midrule
GSPN-T  & [2, 2, 7, 2] & [96, 192, 384, 768] & [4.0, 4.0, 4.0, 4.0] & [L, L, G, G] \\
GSPN-S  & [3, 3, 9, 3] & [108, 216, 432, 864] & [4.0, 4.0, 4.0, 4.0] & [L, L, G, G]  \\
GSPN-B  & [4, 4, 15, 4] & [120, 240, 480, 960] & [4.0, 4.0, 4.0, 4.0] & [L, L, G, G]  \\
\bottomrule
\end{tabular}
}
\end{table*}